\documentclass[runningheads]{llncs}
\usepackage[T1]{fontenc}
\usepackage{graphicx}
\usepackage{booktabs}
\usepackage[misc]{ifsym}

\usepackage{amsmath,amssymb,amsfonts}
\usepackage{float}
\usepackage{xspace} 
\usepackage{subcaption}
\usepackage{tikz}
\usepackage{xcolor}
\usepackage[]{hyperref}
\usepackage{array}
\usetikzlibrary{positioning, shapes.geometric, arrows, fit}
\newcommand{\maxcover}{\textsc{Max-k-Cover}}
\newcommand{\coveragelinearcost}{\textsc{Coverage-Linear-Cost}}
\newcommand{\coveragegraphcost}{\textsc{Coverage-Graph-Cost}}
\newcommand{\teamformation}{\textsc{TeamFormation}}

\newcommand{\task}{\ensuremath{J}\xspace}
\newcommand{\experts}{\ensuremath{\mathcal{X}}\xspace}
\newcommand{\expert}{\ensuremath{X}\xspace}
\newcommand{\skills}{\ensuremath{S}\xspace}
\newcommand{\cov}{\ensuremath{Cov}\xspace}

\newcommand{\costfunction}{\ensuremath{Cost}\xspace}
\newcommand{\costfunctionk}{\ensuremath{Cost_{k}}\xspace}
\newcommand{\costfunctionlinear}{\ensuremath{Cost_{L}}\xspace}
\newcommand{\costfunctiongraph}{\ensuremath{Cost_{G}}\xspace}
\newcommand{\cost}{\ensuremath{\kappa}\xspace}

\newcommand{\assignment}{\ensuremath{\mathbf{x}}\xspace}

\newcommand{\vecy}{\ensuremath{\mathbf{y}}\xspace}
\newcommand{\vecc}{\ensuremath{\mathbf{c}}\xspace}
\newcommand{\vecs}{\ensuremath{\mathbf{s}}\xspace}
\newcommand{\vecb}{\ensuremath{\mathbf{b}}\xspace}
\newcommand{\edges}{\ensuremath{E}\xspace}

\newcommand{\objective}{\ensuremath{F}\xspace}
\newcommand{\opt}{\ensuremath{\mathit{OPT}}\xspace}
\newcommand{\numskills}{\ensuremath{m}\xspace}
\newcommand{\numexperts}{\ensuremath{n}\xspace}
\newcommand{\numtasks}{\ensuremath{t}\xspace}

\newcommand{\penaltymat}{\ensuremath{P}\xspace}
\newcommand{\normobjective}{\ensuremath{\hat{\objective}(\assignment)}\xspace}
\newcommand{\meancov}{\ensuremath{\overline{\cov}}\xspace}
\newcommand{\meansize}{\ensuremath{\overline{z}}\xspace}

\newcommand{\bibsonomy}{\textit{Bbsm}\xspace}
\newcommand{\bibsonomyone}{{\bibsonomy}\textit{-1}\xspace}
\newcommand{\bibsonomytwo}{{\bibsonomy}\textit{-2}\xspace}
\newcommand{\bibsonomythree}{{\bibsonomy}\textit{-3}\xspace}

\newcommand{\imdb}{{\textit{IMDB}}\xspace}
\newcommand{\imdbone}{{{\imdb}\textit{-1}}\xspace}
\newcommand{\imdbtwo}{{{\imdb}\textit{-2}}\xspace}
\newcommand{\imdbthree}{{{\imdb}\textit{-3}}\xspace}

\newcommand{\freelancer}{{\textit{Freelancer}}\xspace}
\newcommand{\freelancerone}{{{\freelancer}\textit{-1}}\xspace}
\newcommand{\freelancertwo}{{{\freelancer}\textit{-2}}\xspace}

\newcommand{\greedy}{\normalfont{\texttt{Greedy}}\xspace}
\newcommand{\qubognn}{\normalfont{\texttt{QUBO-GNN}}\xspace}
\newcommand{\baselinetopk}{\normalfont{\texttt{Top\-k}}\xspace}

\newcommand{\qsolver}{\normalfont{\texttt{Qsolver}}\xspace}
\newcommand{\qubognnRandom}{\normalfont{\texttt{QUBO-GNN-Rand}}\xspace}
\newcommand{\qubognnSimilar}{\normalfont{\texttt{QUBO-GNN-Sim}}\xspace}
\newcommand{\qsolverSimilar}{\normalfont{\texttt{Qsolver-Sim}}\xspace}

\newcommand{\spara}[1]{\smallbreak\noindent{\bf{#1}}}
\newcommand{\mpara}[1]{\medskip\noindent{\bf{#1}}}

\newcommand{\etal}{{et al.}}
\newcommand{\ie}{{i.e.}}

\newcommand*{\belowrulesepcolor}[1]{%
	\noalign{%
		\kern-\belowrulesep
		\begingroup
		\color{#1}%
		\hrule height\belowrulesep
		\endgroup
	}%
}
\newcommand*{\aboverulesepcolor}[1]{%
	\noalign{%
		\begingroup
		\color{#1}%
		\hrule height\aboverulesep
		\endgroup
		\kern-\aboverulesep
	}%
}

\newcommand{\squishlist}{\begin{list}{$\bullet$}
  { \setlength{\itemsep}{0pt}
     \setlength{\parsep}{3pt}
     \setlength{\topsep}{3pt}
     \setlength{\partopsep}{0pt}
     \setlength{\leftmargin}{1.5em}
     \setlength{\labelwidth}{1em}
     \setlength{\labelsep}{0.5em} } }
\newcommand{\squishend}{
  \end{list}  }

\tikzstyle{stage} = [draw, dashed, thick, rounded corners, inner sep=4pt, label={[align=center]above:}]
\tikzstyle{block} = [rectangle, rounded corners=5pt, minimum width=2.2cm, minimum height=0.7cm, text centered, draw=black, thick, fill=orange!40]
\tikzstyle{norm_block} = [rectangle, rounded corners=5pt, minimum width=2.2cm, minimum height=0.7cm, text centered, draw=black, thick, fill=pink!50]
\tikzstyle{output_block} = [trapezium, minimum width=2.2cm, minimum height=0.7cm, text centered, draw=black, thick, fill=green!30]
\tikzstyle{activation} = [rectangle, rounded corners=5pt, minimum width=2.2cm, minimum height=0.7cm, text centered, draw=black, thick, fill=yellow!30]
\tikzstyle{data_skill} = [rectangle, rounded corners=10pt, minimum width=2.2cm, minimum height=0.75cm, text centered, draw=black, thick, fill=gray!50]
\tikzstyle{data_expert} = [rectangle, rounded corners=10pt, minimum width=2.2cm, minimum height=0.75cm, text centered, draw=black, thick, fill=gray!50]
\tikzstyle{arrow} = [thick,->,>=stealth, color=black, line width=0.5mm]
\tikzstyle{arrow1} = [thick,->,>=stealth, color=black, line width=0.5mm]
\tikzstyle{arrow2} = [thick,->,>=stealth, color=black, line width=0.5mm]

\begin{document}

\title{A QUBO Framework for Team Formation}

\titlerunning{A QUBO Framework for Team Formation}

\author{Author information scrubbed for double-blind reviewing}
\author{Karan Vombatkere\inst{1} \Letter \and
Theodoros Lappas\inst{2}\and
Evimaria Terzi\inst{1}}

\authorrunning{K. Vombatkere et al.}
\institute{Boston University \email{\{kvombat,evimaria\}@bu.edu} 
\and Satalia \email{theodoros.lappas@satalia.com}}

\tocauthor{Karan Vombatkere, Theodoros Lappas, Evimaria Terzi}
\toctitle{A QUBO Framework for Team Formation}

\maketitle 

\begin{abstract}
The team formation problem assumes a set of experts and a task, where each expert has a set of skills and the task requires some skills. The objective is to find a set of experts that maximizes coverage of the required skills while simultaneously minimizing the costs associated with the experts. 
Different definitions of cost have traditionally led to distinct problem formulations and algorithmic solutions.
We introduce the unified {\teamformation} formulation that captures all cost definitions for team formation problems that balance task coverage and expert cost. 
Specifically, we formulate three {\teamformation} variants with different cost functions using quadratic unconstrained binary optimization (QUBO), and we evaluate two distinct general-purpose solution methods.
We show that solutions based on the QUBO formulations of {\teamformation} problems are at least as good as those produced by established baselines. 
Furthermore, we show that QUBO-based solutions leveraging graph neural networks can effectively learn representations of experts and skills to enable transfer learning, allowing node embeddings from one problem instance to be efficiently applied to another.

\end{abstract}

\section{Introduction}\label{sec:intro}
The team formation problem is commonly defined as follows: given a set of experts, each possessing a set of skills, and a task that requires specific skills, the goal is to identify a subset of experts best suited to complete the task. A vibrant stream of literature has been dedicated to algorithmic solutions for addressing an ever-expanding universe of variants of this problem~\cite{anagnostopoulos10power,anagnostopoulos2012online,hamidi2023variational,kargar2012efficient,lappas2009finding,nikolakaki20finding,vombatkere2025forming,vombatkere2023balancing}.

The fundamental requirements in most team formation problems is that the selected experts 
maximize the \emph{coverage} of the required skills while minimizing their \emph{cost}.
Existing work on this problem combines these two requirements, by setting one as a constraint and the other as the objective.  
The cost of a team has many different definitions with each leading to a different problem formulation. Common cost functions include a linear sum of individual expert costs or a network-based cost that accounts for the structural connectivity of the selected experts within an underlying social graph.

Inspired by recent work~\cite{nikolakaki21efficient,vombatkere2025forming}, we integrate both the coverage and cost objectives aiming to find a team $\assignment$ for task $J$ such that 
$\lambda \cov (\task \mid \assignment) - \costfunction(\assignment)$
is maximized. We call this general problem 
{\teamformation}. In this formulation, 
\(\lambda\) is a normalization factor that balances the two components of the objective.
This formulation is general and can incorporate direct costs associated with experts or more complex cost functions, e.g., coordination costs.

In this paper, we examine three variants of the {\teamformation} problem resulting from different cost functions, and show that they can be expressed as quadratic unconstrained binary optimization (QUBO) problems. This perspective enables us to frame team formation as an energy minimization problem, drawing parallels with physics-based combinatorial optimization techniques.

\begin{figure}[t]
    \centering
    \includegraphics[width=\textwidth]{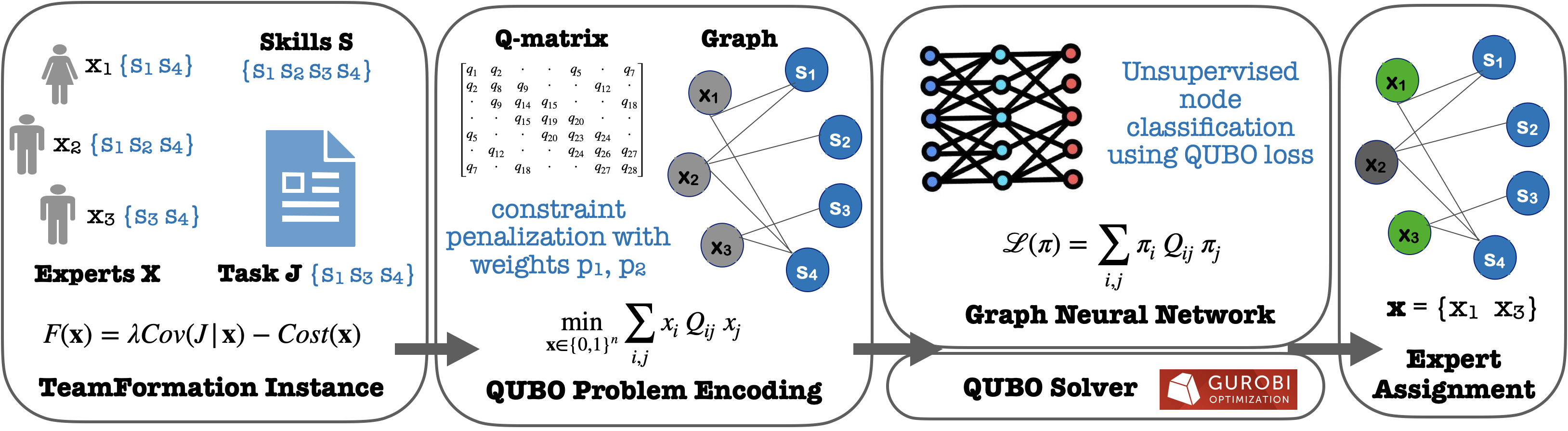}
    \caption{High-level flowchart of our QUBO framework for {\teamformation}.}
    \label{fig:summary-flowchart}
\end{figure}
We explore two classes of solution methods: one using QUBO solvers~\cite{gurobi_optimods} and another leveraging graph neural networks (GNNs)~\cite{scarselli2008graph}. QUBO solvers provide exact or near-optimal solutions. 
However, they operate as black-box solvers that do not provide any insight into the underlying space of experts and skills, and their computational complexity grows significantly with problem size. 

Motivated by these limitations, and inspired by recent work on deep learning for combinatorial optimization problems~\cite{cappart2023combinatorial,schuetz2022combinatorial}, we introduce a GNN-based approach. 
This approach models the problem as an unsupervised node classification task; the classification process assigns each expert a binary decision (selected or not selected in the team) and the GNN learns to classify the experts by optimizing a QUBO-based loss function that corresponds to maximizing the {\teamformation} objective.
Apart from learning good solutions, the embeddings learned by the GNN provide a semantic representation of the expert-skill space, where node proximity reflects relationships between skills and experts. 

To the best of our knowledge, we are the first to provide a unified QUBO-based framework (see Fig.~\ref{fig:summary-flowchart}) for team formation, enabling a consistent algorithmic approach across different {\teamformation} variants. In our experimental evaluation, we utilize real-world datasets from diverse domains, including collaboration networks of artists and scientists, and online labor market data. 
Our results demonstrate that our general algorithms consistently find high-quality solutions, often outperforming combinatorial baselines designed specifically for certain problem variants. Furthermore, our experiments highlight the potential for transfer learning, where GNNs trained on one problem instance can be effectively used to solve related instances with minimal additional computation.

\section{Related Work}\label{sec:related}
Our QUBO-based formulation for the {\teamformation} problem applies to all variants requiring a balance between coverage and cost. 
In this way, our work generalizes a lot of existing work on team formation, relates to work on balancing submodular objectives with other objective functions, and extends ideas from QUBO combinatorial optimization and deep learning.

\spara{Algorithmic Team Formation.} Early work in team formation focused on algorithmic methods to select experts to collectively cover \textit{all} the skills required by a single task, while collaborating effectively within a social network~\cite{kargar2012efficient,lappas2009finding}. Related work considered forming multiple teams of experts to cover the skills of multiple tasks while bounding the workload or coordination cost across experts~\cite{anagnostopoulos10power,anagnostopoulos2012online}. Follow-up works then considered more flexible problem variations that aim to balance partial task coverage with expert cost, maximum workload, and coordination cost. These works primarily employ established algorithmic methods, such as integer programming 
and greedy heuristics~\cite{nikolakaki20finding,nikolakaki21efficient,vombatkere2025forming,vombatkere2023balancing}.

More recent literature has expanded beyond such methods to leverage deep learning for various team-formation variants. For instance, deep neural networks have been used to recommend new teammates to optimally compose high-performance teams~\cite{dashti2022effective,sapienza2019deep}. In another relevant example, a variational bayesian neural architecture was used to learn representations for teams whose members have collaborated in the past, enabling the selection of top-$k$ teams of experts that collectively cover a set of skills~\cite{hamidi2023variational}.

Our {\teamformation} formulation generalizes several prior formulations by incorporating task coverage and a flexible cost definition into a single objective. Furthermore, our GNN-based method is distinct from the deep learning methods used in prior work.

\spara{Submodular Maximization.} The coverage function is monotone and submodular, which is useful within discrete objective functions, as it encodes a natural
diminishing returns property and also comes with an extensive literature on optimization techniques~\cite{feige2011maximizing,feldman2021guess,krause2014submodular}. The greedy algorithm achieves a $1 - 1/e$ approximation for maximizing a nonnegative monotone
submodular function subject to a cardinality constraint~\cite{nemhauser1978analysis}.
There is also work involving maximizing submodular minus modular or linear functions, where no multiplicative approximation guarantees are possible in polynomial time due to potential negativity~\cite{harshaw2019submodular,hochbaum1997approximating}. 

The $\cov()$ function in our {\teamformation} objective is nonnegative monotone submodular, and depending on the definition of $\costfunction()$ used, variants of our general problem relate to balancing submodular and other functions. However, our solution framework is general and it does not rely on the fact that our functions have these properties.

\spara{Combinatorial Optimization and QUBO.}
Many NP-hard combinatorial optimization problems have been formulated as QUBO problems~\cite{glover2019quantum,lucas2014ising}. 
More recently, QUBO has been used as a framework for mapping discrete optimization problems to quantum and classical solvers. Methods for encoding problem constraints, such as unbalanced penalization and slack variable techniques, enable the transformation of constrained combinatorial optimization problems into QUBO~\cite{ayodele2022penalty,montanez2022unbalanced,quintero2022qubo,verma2022penalty}. 

We borrow ideas from prior work to formulate {\teamformation} problems as combinatorial optimization using QUBO, and use the unbalanced penalization technique~\cite{montanez2022unbalanced} to make our formulation more efficient.

\spara{Deep Learning for Combinatorial Optimization.}
Neural combinatorial optimization has gained traction as an alternative to traditional optimization methods, and recent work in reinforcement learning has explored policy-gradient methods and graph-based architectures~\cite{bello2016neural,caramanis2023optimizing}. Neural networks have also been used to learn representations of discrete sets effectively, enhancing the performance of models in tasks involving set-structured data.~\cite{skianis2020rep,zaheer2017deep}.

GNNs have been used to augment existing solvers by identifying smaller sub-problems to reduce the search space for NP-hard problems such as Set Cover~\cite{shafi2023graph}. More closely related to our work, GNNs have been used to solve QUBO-formulated combinatorial optimization problems such as Maximum Independent Set and Maximum Cut, by leveraging their ability to encode graph structures and learn meaningful representations~\cite{schuetz2022combinatorial}.

We extend ideas from the deep learning combinatorial optimization literature to design our GNN architecture to solve the {\teamformation} problem.

\section{Technical Preliminaries}\label{sec:preliminaries}

\subsection{Team Formation}
\spara{Experts, tasks and skills.}
Consider a set of $\numexperts$ experts
$\experts = \{\expert_1,\allowbreak\ldots,\allowbreak\expert_\numexperts\}$, and a single task $\task$.
We assume a set of $\numskills$ skills $\skills$ such that the task $\task$ \emph{requires} a set of skills (i.e., $\task\subseteq\skills$)
and every expert $\expert_i$ \emph{masters} a set of skills (i.e., $\expert_i\subseteq \skills$). 

\spara{Assignments.}
We represent an \emph{assignment} of experts to a task $\task$ using~$\assignment\in\{0,1\}^n$; 
$\assignment(i)=1$ (resp.\ $\assignment(i)=0$) if expert $\expert_i$ is (resp.\ not) assigned to $\task$.
\spara{Task Coverage.}
Given an assignment $\assignment$, we define the \emph{coverage} of task $\task$, denoted by $\cov(\task \mid \assignment)$, as the number of skills required by $\task$ that are covered by the experts assigned to $\task$. That is,
$\cov(\task \mid \assignment) = |(\cup_{i \in \assignment} \expert_i) \cap \task|$, with
$0\leq \cov(\task \mid\assignment)\leq |\task|$. We denote the \emph{size} of $\assignment_i$, i.e., the assignment for task $\task_i$, by $z_i = ||\assignment_i||_1$. This corresponds to the sum of 1-entries in $\assignment_i$.

\spara{Expert Costs.} The cost of an
assignment $\assignment$, denoted by $\costfunction(\assignment)$, encodes the cost of hiring the experts chosen in $\assignment$. Inspired by prior related research, we consider the following established definitions of cost:

\emph{Cardinality cost:} It is often necessary to constrain the \textit{size} of the team, such that the total number of assigned experts is less than or equal to a specified size constraint $k$. This can be encoded as:
    
    $$\costfunctionk(\assignment) = \begin{cases}
        0 & \text{if } \left|(\cup_{i \in \assignment} \expert_i)\right| \leq k \\
        \infty & \text{otherwise}.        
    \end{cases}$$

\emph{Linear cost:} The linear cost is based on ideas first introduced by Nikolakaki {\etal}~\cite{nikolakaki21efficient}. In this case,  each expert $\expert_i$ is associated with a cost $\cost_i$, representing the cost of hiring that expert. The total cost of an assignment $\assignment$ is the sum of costs of the individual experts in the assignment:
\begin{equation*}
    \costfunctionlinear(\assignment) = \sum_{i \in \assignment} \cost_i.
\end{equation*}

\emph{Network coordination cost:} When a set of experts is hired, then there is coordination cost among the experts. We model this by assuming that there is a graph $G = (\experts, \edges)$ between the experts (nodes) and that their pairwise coordination costs are encoded in the weights of the edges between them.
We thus assume that $d(\expert_i,\expert_j): \edges \to \mathbb{R}_{\geq 0}$ 
encodes the coordination cost between two experts. The relevant literature has suggested multiple definitions of coordination cost based on such underlying graphs~\cite{anagnostopoulos2012online,lappas2009finding,vombatkere2025forming}. Inspired by prior work, we define the total coordination cost of an assignment $\assignment$ as the sum of pairwise costs of experts in the assignment:
\begin{equation*}
    \costfunctiongraph(\assignment) = \sum_{(i \in \assignment, j \in \assignment)} d(\expert_i,\expert_j).
\end{equation*}

\subsection{Quadratic Unconstrained Binary Optimization}\label{sec:qubo-preliminaries}
Quadratic unconstrained binary optimization (QUBO) is a mathematical optimization framework used to model combinatorial problems where variables take binary values. For a vector ${\assignment} = (x_1, x_2, \ldots, x_n)$ of binary decision variables (\( x_i \in \{0,1\} \)), the objective function is represented as a quadratic expression of these binary variables:  
\begin{equation}
    \min_{{\assignment} \in \{0,1\}^n} {\assignment}^T Q {\assignment} \;=  \min_{{\assignment} \in \{0,1\}^n} \sum_{i,j}x_i \, Q_{ij} \, x_j,
\end{equation}
where \( Q \) (i.e. the $Q$-matrix) is an \( n \times n \) symmetric matrix, with entries $Q_{ij}$. The $Q$-matrix encodes problem-specific interactions between variables. QUBO is an NP-hard optimization problem~\cite{mehta2022hardness}.

\mpara{Solvers.} Classical solvers, such as Gurobi's QUBO optimizer and CPLEX, use mixed-integer programming (MIP), branch-and-bound, and specialized heuristic methods to find optimal or near-optimal solutions to QUBO problems~\cite{gurobi_optimods}.

\mpara{Linear Programs as QUBO.} A linear program (LP) with binary variables $\assignment$ can be represented as QUBO by reformulating equality constraints using quadratic penalty terms~\cite{glover2019quantum,quintero2022qubo}. 
Consider an LP of the form \(\min {\vecc}^T \assignment \) subject to equality constraints \( A \assignment = \vecb \), where $\assignment$ is any length-$n$ binary vector, $A$ is a $(m \times n)$ matrix and $\vecb$ is a length-$m$ vector. Denoting $C = \text{diag}({\vecc})$, and for an appropriate scalar penalty $p$ we have the following equivalence:
\begin{align*}
    \min_{\assignment} {\vecc}^T {\assignment}\; (\text{s.t. } A \assignment = \vecb) &= \min_{\assignment} {\assignment}^T C {\assignment} + p(A\assignment - \vecb)^T (A\assignment - \vecb) \\
    &= \min_{\assignment} \assignment^T Q \assignment  + p\vecb^T\vecb.
\end{align*}
The optimal solution to the LP 
\(\min_{\assignment} {\vecc}^T \assignment \) subject to \( A \assignment = \vecb \) corresponds to the optimal solution to $\min_{\assignment} \assignment^T Q \assignment$, where $Q = C + p(A^T A) - 2p\; \text{diag}(A^T \vecb)$ is the $Q$-matrix of the QUBO encoding, and we dropped the additive constant $p\vecb^T\vecb$.

\mpara{Unbalanced Penalization.} To transform an LP with inequality constraints, typically slack variables are introduced
as follows: 
given a constraint \( \sum_i a_{ij} x_i \leq b_j \), where $a_i, b_j \in \mathbb{Z}$ for every $j=\{1,\ldots ,m\}$, a non-negative slack variable encoded as a sum of binary variables \( \hat{s} = \sum_k 2^k s_k \) (where \( s_k \in \{0,1\} \)), is added so the constraint becomes \( \sum_i a_i x_i + \hat{s} = b_j \). The reformulated equality is then enforced in the objective function using a quadratic penalty term \( p (\sum_i a_i x_i + \sum_k 2^k s_k - b_j)^2 \), where \( p \) is a sufficiently large penalty coefficient. 

The primary drawback of slack variables is the increase in dimensionality of the LP -- and the size of the $Q$-matrix -- by $\log \lceil b_j - \sum_i a_i x_i\rceil$ for each inequality constraint.
Consequently, for the problems in this paper, we eliminate the need for slack variables by incorporating unbalanced penalization~\cite{montanez2022unbalanced}.
This technique encodes an asymmetric penalty function (directly into the QUBO objective) which is small when a constraint is satisfied and increases significantly when violated, without increasing the problem's dimensionality.

We provide all mathematical details to use unbalanced penalization to formulate team formation LPs into QUBO in Section~\ref{sec:qubo-tf}. 

\section{QUBO Framework for Team Formation}\label{sec:qubo-tf}
 In this section we introduce the general {\teamformation} problem, and detail three variants, which we then formulate using QUBO.

\subsection{The {\teamformation} Problem}
Given a set of experts $\experts$, and a task $\task$, we define the general {\teamformation} problem as follows: find an assignment $\assignment$ that \textit{maximizes} the objective
\begin{equation}\label{eq:general-problem}
    \objective(\assignment) = \lambda \cov(\task \mid \assignment)  - \costfunction (\assignment).
\end{equation}
The above function balances the coverage of task $\task$ achieved by a specific team with the cost of the team.
Parameter $\lambda$ is application dependent and can be used to tune the importance of the two components of the objective.

We now define three instantiations of the {\teamformation} problem,
which have different cost functions. We express each of these problems using constrained linear programming and apply the unbalanced penalization technique (see Sec.~\ref{sec:qubo-preliminaries}) to construct the corresponding $Q$-matrix. 

Throughout this section we use the vector $\vecy = \vecs \; || \; \assignment$ which represents the solution to our problems. We call $\vecy$ the \emph{solution vector}. This vector is of size $(\numskills + \numexperts)$ and is the concatenation of $\vecs$ and $\assignment$, where $\vecs$ is a binary vector that 
encodes whether a skill $i$ is covered (resp.\ not covered) by
$\vecs$ when $s_i=1$ (resp.\ $s_i=0$). We also use 
the $(\numexperts \times \numskills)$ skill-membership matrix $E$ such that $E(i,j)=1$ (resp. $0$) if expert $i$ has (resp.\ not) skill $j$. 

Due to space constraints, we omit several mathematical details and refer the reader to the supplementary material for derivations of the QUBO formulations.

\subsection{{\maxcover}}
\begin{problem}[{\maxcover}]{\label{problem:maxkcover}}
Given a set of $\numexperts$ experts
$\experts = \{\expert_1,\allowbreak\ldots,\allowbreak\expert_\numexperts\}$, a task $\task$, and a cardinality constraint $k$, find an assignment $\assignment$ of experts such that the following is maximized:
\begin{equation}\label{eq:objective}
    \objective(\assignment) = \lambda \cov(\task \mid\assignment) - \costfunctionk(\assignment).
\end{equation}
\end{problem}

\spara{QUBO Formulation Sketch.} Let $\vecy = \vecs \; || \; \assignment$, be the $(\numskills + \numexperts)$-size solution vector we described above. Now let $\vecc$
be another $(\numskills+\numexperts)$ vector such that
$c_i = \lambda$ if $i \leq \numskills$ \textit{and} skill $i \in \task$, and $c_i = 0$ otherwise.
Then, the  {\maxcover}  problem can be expressed by the following linear program:
\begin{align*}
    \text{maximize}~~  & {\vecc}^T {\vecy},\\
    \text{such that}~~ & \sum_{i=1}^{\numexperts} x_{i} \leq k \\
    &s_j - \sum_{i=1}^{\numexperts} E(i, j) \cdot x_i \leq 0 \quad \text{ for all } 1 \leq j \leq \numskills , \text{ and}\\
    & s_i, x_{i}\in \{0, 1\}.
\end{align*}
We derive penalty matrices $\penaltymat_k$ and $\penaltymat_C$ corresponding to the LP constraints. 
Then the $(\numskills + \numexperts) \times (\numskills + \numexperts)$ square matrix 
$Q = -\text{diag}({\vecc}) - \penaltymat_k + \penaltymat_C$
provides a QUBO formulation of {\maxcover}, where minimizing ${\vecy}^T Q {\vecy}$ corresponds to maximizing $\objective(\assignment) = \lambda \cov(\task \mid \assignment) - \costfunctionk(\assignment)$.

\subsection{{\coveragelinearcost}}
\begin{problem}[{\coveragelinearcost}]{\label{problem:coveragelinearcost}}
Given a set of $\numexperts$ experts
$\experts = \{\expert_1,\allowbreak\ldots,\allowbreak\expert_\numexperts\}$
with their corresponding individual costs $\{\cost_1, \allowbreak\ldots,\allowbreak\cost_\numexperts\}$, 
and a task $\task$,
find an assignment $\assignment$ of experts such that the following is maximized:
\begin{equation}\label{eq:objective}
    \objective(\assignment) = \lambda \cov(\task \mid \assignment)  - \costfunctionlinear (\assignment).
\end{equation}
\end{problem} 

\spara{QUBO Formulation Sketch.} 
Let $\vecy = \vecs \; || \; \assignment$, be the $(\numskills + \numexperts)$-size solution vector we described above. Now let $\vecc$
be another $(\numskills+\numexperts)$ vector 
such that $c_i = \lambda$ if $i \leq \numskills$ \textit{and} skill $i \in \task$, $c_i = -\cost_{i-\numskills}$ if $i > \numskills$; recall that 
$\cost_{i}$ is the cost of hiring expert $i$ (see Sec.~\ref{sec:preliminaries}). 
Then {\coveragelinearcost} can be expressed as:
\begin{align*}
    \text{maximize}~~  & {\vecc}^T {\vecy},\\
    \text{such that}~~ & s_j - \sum_{i=1}^{\numexperts} E(i, j) \cdot x_i \leq 0 \quad \text{ for all } 1 \leq j \leq \numskills , \text{ and}\\
    & s_i, x_{i}\in \{0, 1\}.
\end{align*}
We create penalty matrices $\penaltymat_1$ and $\penaltymat_2$ to capture the constraints in the LP. 
Then, the $(\numskills + \numexperts) \times (\numskills + \numexperts)$ square matrix 
$Q = -\text{diag}({\vecc}) - \penaltymat_1 + \penaltymat_2$
has the property that minimizing ${\vecy}^T Q {\vecy}$ corresponds to maximizing $\objective(\assignment) = \lambda \cov(\task \mid \assignment)  - \costfunctionlinear (\assignment)$.

\subsection{{\coveragegraphcost}}
\begin{problem}[{\coveragegraphcost}]{\label{problem:coveragegraphcost}}
Given a set of $\numexperts$ experts
$\experts = \{\expert_1,\allowbreak\ldots,\allowbreak\expert_\numexperts\}$ with a corresponding distance function $d(\cdot,\cdot)$ between any pair of experts, 
and a task $\task$,
find an assignment $\assignment$ of experts such that we maximize:
\begin{equation}\label{eq:objective}
    \objective(\assignment) = \lambda \cov(\task \mid \assignment)  - \costfunctiongraph (\assignment).
\end{equation}
\end{problem} 

\spara{QUBO Formulation Sketch.} We consider the following constrained linear program that encodes the {\coveragegraphcost} problem:
\begin{align*}
    \text{maximize}~~  & \lambda \cdot \sum_{i=1}^{n} s_i - \sum_{(i,j)} d(i,j) \cdot (x_i x_j)\\
    \text{such that}~~ &s_j - \sum_{i=1}^{\numexperts} E(i, j) \cdot x_i \leq 0 \quad \text{ for all } 1 \leq j \leq \numskills , \text{ and}\\
    & s_i, x_{i}\in \{0, 1\}.
\end{align*}
For the QUBO formulation we need the solution vector $\vecy$, we defined above. We also need 
the $(\numskills + \numexperts)$ vector ${\vecc} = (c_{1}, \ldots, c_{(\numskills + \numexperts)})$, such that $c_i = \lambda$ if $i \leq \numskills$ \textit{and} skill $i \in \task$, and $c_i = 0$ otherwise. Then we compute the $(\numexperts \times \numexperts)$ matrix $D$ of pairwise distances such that $D(i,j) = d(\expert_i, \expert_j)$ and add it to the lower-right $(\numexperts \times \numexperts)$ submatrix of $\text{diag}({\vecc})$ to obtain $\hat{D} = \text{diag}({\vecc}) + 
\begin{bmatrix}
    \mathbf{0}_{m \times m} & \mathbf{0}_{m \times n} \\
    \mathbf{0}_{n \times m} & \;D_{n \times n}
 \end{bmatrix}$
Now, $\objective(\assignment) = \vecy^T \hat{D} \vecy$ encodes the {\coveragegraphcost} objective. We create penalty matrices $\penaltymat_1, \penaltymat_2$ to capture the LP constraints; the $(\numskills + \numexperts) \times (\numskills + \numexperts)$ square matrix 
$Q = -\hat{D} - \penaltymat_1 + \penaltymat_2$ provides a complete QUBO formulation of {\coveragegraphcost}; that is, minimizing ${\vecy}^T Q {\vecy}$ corresponds to maximizing $\objective(\assignment) = \lambda \cov(\task \mid \assignment)  - \costfunctiongraph(\assignment)$.\\

All three {\teamformation} problem variants
are hard to solve and approximation and heuristic algorithms exist in the literature~\cite{harshaw2019submodular,krause2014submodular,nikolakaki21efficient}. 

\section{Solving {\teamformation} Problems}\label{sec:gnn}
In this section, we describe two different general-purpose methods that leverage the QUBO formulation to solve {\teamformation} problems.

\subsection{QUBO Solver}
We use a QUBO solver implemented by Gurobi~\cite{gurobi_optimods}. The solver takes the $Q$-matrix corresponding to a QUBO problem as input, 
and applies mixed-integer programming methods with specialized heuristics to solve the QUBO instance. 
We use Gurobi's QUBO solver with the $Q$-matrix corresponding to the {\teamformation} problems, and refer to this method as {\qsolver}.

\subsection{Graph Neural Networks}
Combinatorial optimization problems are formulated as QUBO~\cite{schuetz2022combinatorial} and represented as a graph \( G = (V, E) \), where each vertex \( i \in V \) corresponds to a binary decision variable \( y_i \in \{0,1\} \). The objective function is defined by a Hamiltonian \( \mathbb{H}(\vecy) \), which represents the system’s energy. 
The binary state \( y_i \) is relaxed into a continuous representation \( \pi_i \in [0,1] \), allowing gradient-based optimization to be applied. The architecture employs multiple layers of message-passing neural networks to iteratively update node representations. At each layer \( l \), the hidden state \( \pi_i^{(l)} \) of node \( i \) is updated based on its current state and information aggregated from its neighboring nodes \( \mathcal{N}(i) \):
$\pi_i^{(l+1)} = \sigma \left( W^{(l)} \pi_i^{(l)} + \sum_{j \in \mathcal{N}(i)} W^{(l)} \pi_j^{(l)} + {\bf w_0}^{(l)} \right)$
where \( W^{(l)} \) and \({\bf w_0}^{(l)} \) are the weight matrix and bias vector for layer \( l \), and \( \sigma \) is a nonlinear activation function. The loss function 
is based on the relaxed Hamiltonian \( \mathbb{H}(\bf \pi) \), such that the network is trained to minimize the energy. After training, the continuous node states \( \pi_i \) are projected back to binary \( y_i \), yielding a feasible solution to the original combinatorial optimization problem.

\mpara{GNNs for {\teamformation}.} We perform unsupervised node classification using a GNN to solve the QUBO formulation corresponding to {\teamformation}. 
Given the $Q$ matrix that encodes a problem, the goal is to find the $(\numskills + \numexperts)$-size solution vector $\vecy = \vecs || \assignment$ that minimizes ${\vecy}^T Q {\vecy}$, with 
$\assignment = (y_{m+1}, \ldots, y_{m + n})$ being the desired solution assignment to the {\teamformation} problem. 

\spara{Graph Creation.} We create a graph $G = (V, E)$, where each vertex \(i \in V \) corresponds to a binary decision variable \(y_i \in \{0,1\}\); vertices $(1, \ldots, \numskills)$ correspond to the set of all skills, and vertices $(\numskills+1, \ldots, \numskills+\numexperts)$ correspond to the experts in the {\teamformation} problem instance. 
For every skill each expert has, we create an unweighted edge in $G$ between the corresponding expert and skill vertices, i.e. $E = \{(i,j): s_i \in \expert_j\}$. For {\coveragegraphcost}, we add weighted edges between expert vertices to encode the pairwise network coordination costs.

\spara{Loss Function and Regularization.} 
Since ${\vecy}^T Q {\vecy}$ is not differentiable and cannot be used as such within the GNN training process, we follow the approach of Schuetz {\etal}~\cite{schuetz2022combinatorial} to relax each binary variable \(y_i \in \{0,1\}\) such that $y_i \rightarrow \pi_i \in [0,1]$, where these $\pi_i$ can be viewed as selection probabilities, i.e. small $\pi_i$ implies $y_i$ is not selected, and large $\pi_i$ implies $y_i$ is selected. 
We then generate the following differentiable loss function used for backpropagation:
\begin{equation*}
 \mathcal{L}(\pi) = \sum_{i,j}\pi_i \; Q_{ij} \; \pi_j + \alpha \cdot \sum_i \pi_i \; (1 - \pi_i).
\end{equation*}
We include the regularization term $\alpha \cdot \sum_i \pi_i \; (1 - \pi_i)$ to encourage the GNN to converge to binary solutions, where $\alpha$ is a tunable hyperparameter.

We randomly initialize node embeddings for each of the expert and skill nodes, where the dimension of the embeddings is given by the hyperparameter $d_0$. We denote the set of $(\numskills + \numexperts)$ embeddings by $H^{(0)} = H_{\skills}^{(0)}\; || \; H_{\experts}^{(0)}$, where $||$ represents concatenation of the $\numskills$ skill embeddings and $\numexperts$ expert embeddings.

\spara{Graph Convolution.} Vertices in $G$ represent skills \emph{and} experts, and thus we have two different types of edges: between experts and skills, and between two experts.
To ensure message-passing during GNN training occurs over valid edge types, we adopt a two-layer (heterogeneous) graph convolution network (GCN) architecture, with forward propagation given by $H^{(1)} = \sigma_1 \left( \sum_{r \in \mathcal{R}} \Theta_r^0 H^{(0)} \right)$ and $H^{(2)} = \sigma_2 \left( \sum_{r \in \mathcal{R}} \Theta_r^1 H^{(1)} \right)$,
where \( \mathcal{R} \) is the set of different edge types. 
\( H^{(0)} \) represents the input node embeddings of size $d_{0}$, and \( H^{(1)} \) and \( H^{(2)} \) are the hidden and output layer representations of sizes $d_h$ and $(\numskills + \numexperts)$, respectively. \( \Theta_r^0 \) and \( \Theta_r^1 \) are trainable weight matrices specific to \( r \), allowing the GNN to learn different transformations per edge type; \( \sigma_1, \sigma_2 \) are non-linear activation functions, applied element-wise; we use ReLU for \( \sigma_1 \) and a sigmoid for \( \sigma_2 \).

We add batch normalization after the first graph convolutional layer to normalize activations and stabilize training. We also introduce dropout after the ReLU activation by randomly setting $p_d$ fraction of neurons in the GNN to zero.

\begin{figure}[t]
\centering
\resizebox{0.71\textwidth}{!}
    {\small
\begin{tikzpicture}[node distance=0.4]
    \node (skillEmbed) [data_skill] {$H_{\skills}^{(0)}$ Skill Embeddings ($d_0$)};
    \node (expertEmbed) [data_expert, right=of skillEmbed, xshift=0.2cm] {$H_{\experts}^{(0)}$ Expert Embeddings ($d_h$) };

    \node (conv1) [block, below=of skillEmbed, xshift=2cm] {$H^{(1)}$ Graph Convolution ($d_0 \times d_{h}$)};
    \node (norm1) [norm_block, below=of conv1] {Batch Normalization};
    \node (relu1) [activation, below=of norm1] {$\sigma_1$ ReLU};
    \node (dropout1) [norm_block, below=of relu1] {Dropout ($p_d$)};

    \node [stage, fit=(conv1) (dropout1), label={[align=center]left:\textbf{GCN Layer 1}}] {};

    \node (conv2) [block, below=of dropout1] {$H^{(2)}$ Graph Convolution ($d_{h} \times (\numskills + \numexperts)$)};
    \node (sigmoid) [activation, below=of conv2] {$\sigma_2$ Sigmoid};

    \node [stage, fit=(conv2) (sigmoid), label={[align=center]left:\textbf{GCN Layer 2}}] {};

    \node (rounding) [norm_block, below=of sigmoid] {Concatenate \& Round};

    \node (output) [output_block, below=of rounding] {Output: ${\vecy} \in \{0,1\}^{(\numskills + \numexperts)}$};
    \draw [arrow1] (skillEmbed) -- (conv1);
    \draw [arrow1] (conv1) ++(-0.5cm, -0.33) -- (1.5cm, -1.95cm);
    \draw [arrow1] (norm1) ++(-0.5cm, -0.33) -- (1.5cm, -3.1cm);
    \draw [arrow1] (relu1) ++(-0.5cm, -0.33) -- (1.5cm, -4.2cm);
    \draw [arrow1] (dropout1) ++(-0.5cm, -0.33) -- (1.5cm, -5.35cm);
    \draw [arrow1] (conv2) ++(-0.5cm, -0.33) -- (1.5cm, -6.45cm);
    \draw [arrow1] (sigmoid) ++(-0.5cm, -0.33) -- (1.5cm, -7.6cm);

    \draw [arrow2] (expertEmbed) -- (conv1);
    \draw [arrow2] (conv1) ++(0.5cm, -0.33) -- (2.5cm, -1.95cm);
    \draw [arrow2] (norm1) ++(0.5cm, -0.33) -- (2.5cm, -3.1cm);
    \draw [arrow2] (relu1) ++(0.5cm, -0.33) -- (2.5cm, -4.2cm);
    \draw [arrow2] (dropout1) ++(0.5cm, -0.33) -- (2.5cm, -5.35cm);
    \draw [arrow2] (conv2) ++(0.5cm, -0.33) -- (2.5cm, -6.45cm);
    \draw [arrow2] (sigmoid) ++(0.5cm, -0.33) -- (2.5cm, -7.6cm);

    \draw [arrow] (rounding) -- (output);

    \node[draw, fill=blue!10, rounded corners, thick, align=center, font=\large] at (6.3,-3.3)
    {Unsupervised training \\
    using gradient descent \\
    with QUBO Loss\\
    $\mathcal{L}(\vecy) = \sum_{i,j} \pi_i \; Q_{ij} \; \pi_j + $\\
    $\quad \quad \quad \alpha \cdot \sum_i \pi_i \; (1 - \pi_i)$};    
\end{tikzpicture}
}
\caption{{\qubognn} model architecture for solving {\teamformation} problems.}
\label{fig:gnn_flowchart}
\end{figure}

We call our method {\qubognn} and visualize the model architecture in Figure~\ref{fig:gnn_flowchart}. {\qubognn} is parametrized by several hyperparameters; Table~\ref{tab:hyperparams} provides a summary of the hyperparameters of the {\qubognn} model, and heuristic ranges of values to grid-search. 
\begin{table}[th]
    \centering
    \caption{Description of {\qubognn} model parameters.}
    \label{tab:hyperparams}
    \resizebox{0.9\textwidth}{!}
    {\small
    \begin{tabular}{p{2.5cm} p{4cm} l}
        \toprule
        \textbf{Parameter} & \textbf{Description} & \textbf{Heuristic range} \\
        \midrule
        $p_1$ & QUBO penalty 1 & $[10^{-1}, 10^2]$\\
        $p_2$ & QUBO penalty 2 & $[10^{-1}, 10^2]$\\
        $\lambda$ & Normalizing coefficient & $[1, 10^2]$\\
        $d_0$ &  Size of node embeddings & $[(\numskills + \numexperts)^{1/2}, (\numskills + \numexperts)/2]$\\
        $d_{h}$  &  Size of hidden layer & $[(\numskills + \numexperts)^{1/2}, (\numskills + \numexperts)/2]$\\
        $p_d$ &  Dropout probability & $[0.1, 0.3]$\\
        $\alpha$ & Binary regularization weight & $[1, 10]$\\
        $\beta$ & Learning rate & $[10^{-4}, 10^{-2}]$\\
        \bottomrule
    \end{tabular}
    }
\end{table}
The model hyperparameters $d_0, d_h, p_d, \alpha$ and $\beta$ can be set heuristically or optimized in an outer-loop using grid-search. 

Capturing problem constraints effectively in a QUBO formulation requires the selection of suitable scalar penalties $p_1, p_2$. In practice, we observed for our problems that the unbalanced penalization scheme yields good solutions for a wide range of values of $p_1, p_2$. However, to enable convergence to better near-optimal solutions we implement a grid search for $p_1, p_2$ over the range of heuristic values shown in Table~\ref{tab:hyperparams}.

\spara{Projection Rounding and Output.} At the end of unsupervised training, the $\sigma_2$ sigmoid activation layer outputs probabilities $\pi_i$ associated with each node which we can view as soft assignments. We apply a simple rounding scheme: $y_i = \text{int}(\pi_i)$ to project these probabilities $\pi_i$ back to binary assignments \(y_i \in \{0,1\}\). 

\section{Experimental Analysis}\label{sec:experiments}
\subsection{Experimental Setup}
\spara{Datasets.}
We evaluate our methods on several real-world datasets also used in past team formation papers: {\freelancer}, {\imdb}, {\bibsonomy}~\cite{anagnostopoulos10power,nikolakaki20finding,nikolakaki21efficient,vombatkere2023balancing}. We follow the method of 
\cite{anagnostopoulos2012online} and create social graphs with expert coordination costs for our
datasets.
We provide summary statistics of the datasets in Table~\ref{tab:summarystats}. 
Detailed descriptions and pre-processing steps of each dataset are available in the supplementary material.

\begin{table}[h]
    \centering
    \caption{Summary statistics of our datasets.}
    \label{tab:summarystats}
    \resizebox{0.82\textwidth}{!}
    {\small
    \begin{tabular}{lccccccc}
    \toprule
    Dataset & Experts & Tasks & Skills & \multicolumn{1}{c}{Skills/} & \multicolumn{1}{c}{Skills/} & \multicolumn{1}{c}{Average}  & \multicolumn{1}{c}{Average} \\
     &  &  &  & \multicolumn{1}{c}{expert} & \multicolumn{1}{c}{task} & \multicolumn{1}{c}{path length} &  \multicolumn{1}{c}{degree} \\
    \midrule
        {\freelancerone}&50&250&50&2.2&4.3 &2.6 & 4.5\\
        {\freelancertwo}&150&250&50&2.2&4.4 &2.4 & 10.4\\
        {\imdbone}&200 &300 &23 &3.3 &5.0 &3.0 & 0.4 \\
        {\imdbtwo}&400 &300 &23 &3.8 &5.3 &7.1 & 0.9\\
        {\imdbthree}&1000 &300 &25 &4.5 &5.2 &6.2 & 2.3\\
        {\bibsonomyone}&250 &300 &75 &12.5 &5.5 & 5.9 & 1.9\\
        {\bibsonomytwo}&500 &300 &75 &13.0 &5.5 &2.6 & 9.4\\
        {\bibsonomythree}&1000 &300 &75 &13.1 &5.5 &2.6 & 13.3\\
        \bottomrule
    \end{tabular}
    }
\end{table}

\spara{Baselines.} 
For each of the {\teamformation} variants, we evaluate the performance of {\qubognn} and {\qsolver} against 
some problem-specific baselines, which have the same principles across problem variants. We describe those below.

\spara{{\greedy}:} For {\maxcover} 
the {\greedy} baseline iteratively picks the expert with the maximum
marginal skill coverage. 
For {\coveragelinearcost}, {\greedy} implements 
the {\tt Cost-Scaled Greedy} algorithm introduced by Nikolakaki {\etal}~\cite{nikolakaki21efficient}. For the {\coveragegraphcost} problem,
{\greedy} picks the expert that maximizes the ratio
of coverage over coordination cost  at each iteration.

\spara{{\baselinetopk}:} 
This is an objective-agnostic algorithm that ranks the experts based on their Jaccard similarity with the input task and then picks the top-$k$ most similar experts, where $k$ is determined by the size of the {\greedy} (or {\qsolver}) solution.
\spara{Implementation Details.}
We used single-process implementations on a 14-core 2.4 GHz Intel Xeon E5-2680 processor for all our experiments. We implement our {\qubognn} architecture in Python using PyTorch~\cite{paszke2019pytorch} and Deep Graph Library\cite{wang2019deep}, and fine-tune model hyperparameters using grid search. 
For each dataset, we train separate {\qubognn} models for up to 100 different tasks.
For the normalizing coefficient we set $\lambda = 50$, which yields a reasonable balance between weighting coverage and cost for our {\teamformation} variants. To aid reproducibility, we report the full set of model parameters used in the supplement, and make our code~\footnote{https://github.com/kvombatkere/Team-Formation-QUBO} available online. 

\begin{figure}[ht]
    \centering
    \begin{subfigure}{\textwidth}
        \centering
        \includegraphics[width=\textwidth]{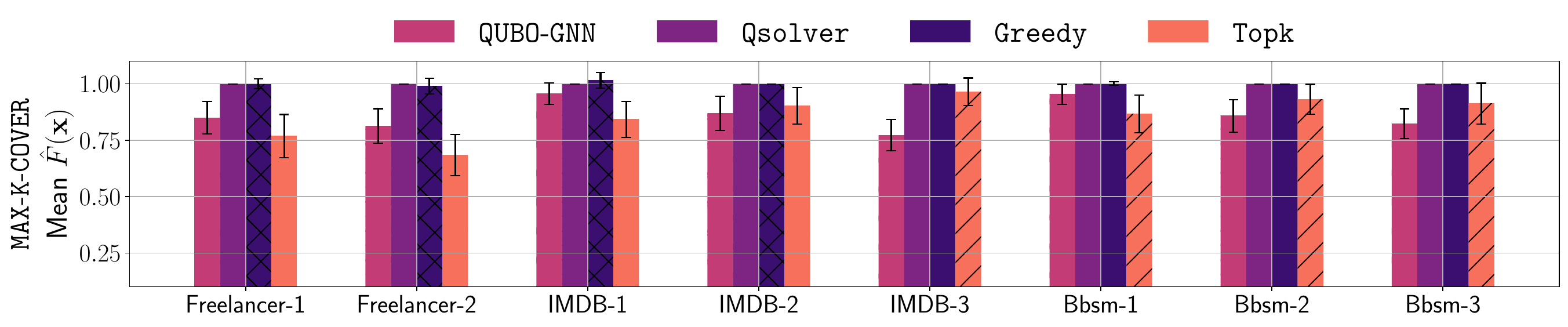}
        \label{fig:maxcover-barplot}
    \end{subfigure}
    \begin{subfigure}{\textwidth}
        \centering
        \includegraphics[width=\textwidth]{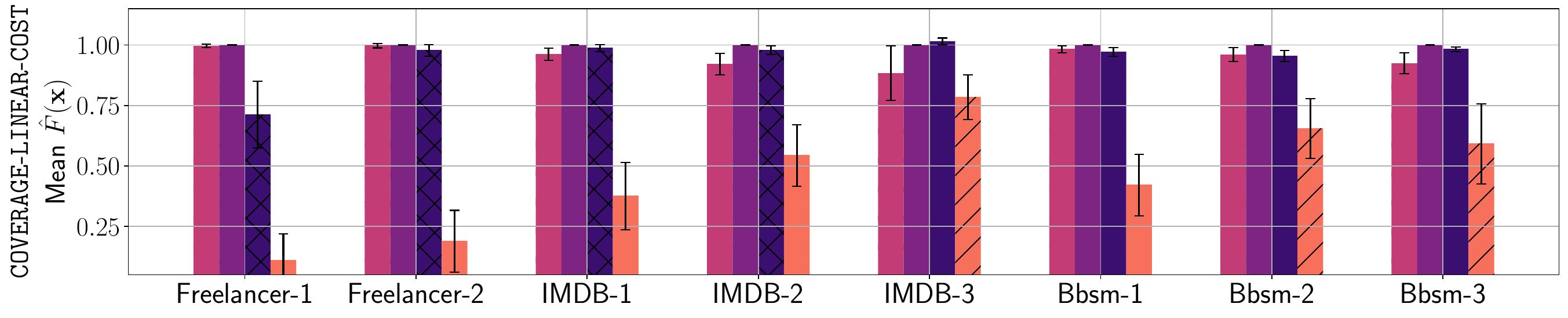}
        \label{fig:covcost-barplot}
    \end{subfigure}    
    \begin{subfigure}{\textwidth}
        \centering
        \includegraphics[width=\textwidth]{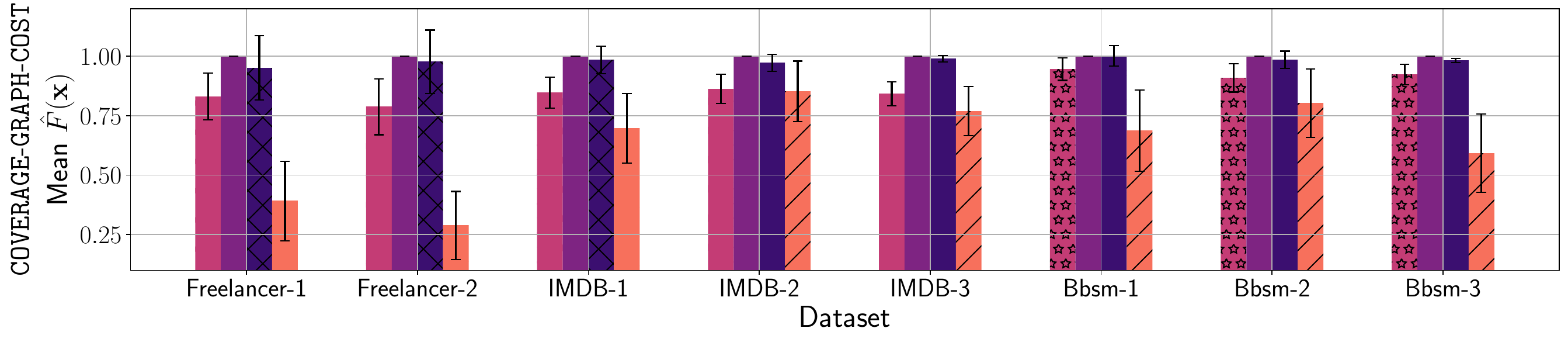}
        \label{fig:sub3}
    \end{subfigure}
    \caption{Bar plots showing the mean {\qsolver}-normalized objective, $\normobjective$ of {\qubognn}, {\qsolver}, {\greedy} and {\baselinetopk}, across all training task instances for all three {\teamformation} variants.}
    \label{fig:objective-barplots}
\end{figure}



\subsection{Quantitative Comparison}\label{sec:quant-comparison}
We evaluate our algorithms against the baselines with respect to our overall objective (and the corresponding coverage, size and cost).
Due to space constraints, we only show detailed results for {\coveragelinearcost}, and provide experimental results for {\maxcover} and {\coveragegraphcost} in the supplement. Note that the general experimental patterns observed were similar across all three {\teamformation} variants.

We observe that {\qsolver} has the best performance for all datasets, and consequently analyze the objective of our methods by first normalizing by the corresponding {\qsolver} objective and then taking the mean across all training tasks. We denote the normalized objective by $\normobjective$. 

\spara{Aggregate Performance Evaluation.} 
Figure~\ref{fig:objective-barplots} presents the mean $\normobjective$ across all training tasks returned by {\qubognn}, {\qsolver}, {\greedy}, and {\baselinetopk} across our datasets. We observe that {\qsolver} consistently achieves the highest normalized mean objective values (i.e., values equal to $1$): it outperforms the other methods across all datasets for all three {\teamformation} variants.  
We observe that {\greedy} performs slightly worse than {\qsolver}, and {\baselinetopk} consistently has the lowest $\normobjective$. For most datasets, {\qubognn} achieves solutions with objective values that are comparable (but slightly worse) than {\qsolver}. Overall, this is expected as {\qsolver} finds the optimal solution for the same problem that {\qubognn}tries to solve. Moreover, the success of both QUBO-based algorithmic solutions demonstrate that our QUBO formulation is appropriate for solving the original team formation problem.

\begin{table}[ht]
    \caption{Mean task coverage, $\meancov$ and solution size, $\meansize$ of {\qubognn}, {\qsolver} and {\greedy} across all training task instances for {\coveragelinearcost}.}
    \centering
    \label{tab:CovCost-Train}
    \resizebox{0.92\textwidth}{!}
    {\small
    \begin{tabular}{l *{3}{>{\centering\arraybackslash}m{1.5cm}} *{3}{>{\centering\arraybackslash}m{1.5cm}}}
    \toprule
    Dataset & \multicolumn{3}{c}{Mean Task Coverage, $\meancov$} & \multicolumn{3}{c}{Mean Solution Size, $\meansize$} \\
    \cmidrule(r){2-4} \cmidrule(l){5-7}
    & {\qubognn} & {\qsolver} & {\greedy} & {\qubognn} & {\qsolver} & {\greedy} \\
    \midrule
    {\freelancerone} & 0.88 & 0.89 & 0.48 & 2.8 & 2.9 & 1.4 \\
    {\freelancertwo} & 0.98 & 0.98 & 0.92 & 3.2 & 3.2 & 2.9 \\
    {\imdbone}     & 0.99 & 1.00 & 0.99 & 2.3 & 2.4 & 2.1 \\
    {\imdbtwo}    & 0.98 & 1.00 & 1.00 & 2.5 & 2.3 & 1.8 \\
    {\imdbthree}     & 0.88 & 1.00 & 1.00 & 2.5 & 3.2 & 1.2 \\
    {\bibsonomyone}    & 1.00 & 1.00 & 1.00 & 3.1 & 2.7 & 2.0 \\
    {\bibsonomytwo}    & 0.97 & 1.00 & 1.00 & 2.8 & 2.6 & 1.6 \\
    {\bibsonomythree}   & 0.98 & 1.00 & 1.00 & 1.8 & 4.2 & 1.7 \\
    \bottomrule
    \end{tabular}
    }
\end{table}

We use $\meancov = \frac{1}{\numtasks}\sum_{i=1}^t \cov(\task_i | \assignment)$ to denote the mean coverage, and $\meansize = \frac{1}{\numtasks}\sum_{i=1}^t z_i$ to denote the mean solution size, across training tasks $\task_1, \ldots, \task_\numtasks$.
We observe from Table~\ref{tab:CovCost-Train} that all three methods find solutions yielding high coverages for {\imdb} and {\bibsonomy}.
However, {\qubognn} and {\qsolver} often find assignments with a larger solution size (and larger cost) than {\greedy}. These assignments -- particularly for {\freelancer} -- lead to higher coverages resulting in superior objective values.
This tradeoff highlights the ability of the QUBO formulation to balance cost and team effectiveness better than greedy approaches. 
Finally, even though {\greedy} was the fastest algorithm in terms of running time, {\qubognn} and {\qsolver} converged to good solutions within a few seconds, even for the largest datasets (i.e. {\imdbthree} and {\bibsonomythree}).

\spara{Individual Task Evaluation.}
\begin{figure}[t]
    \centering
    \includegraphics[width=0.95\textwidth]{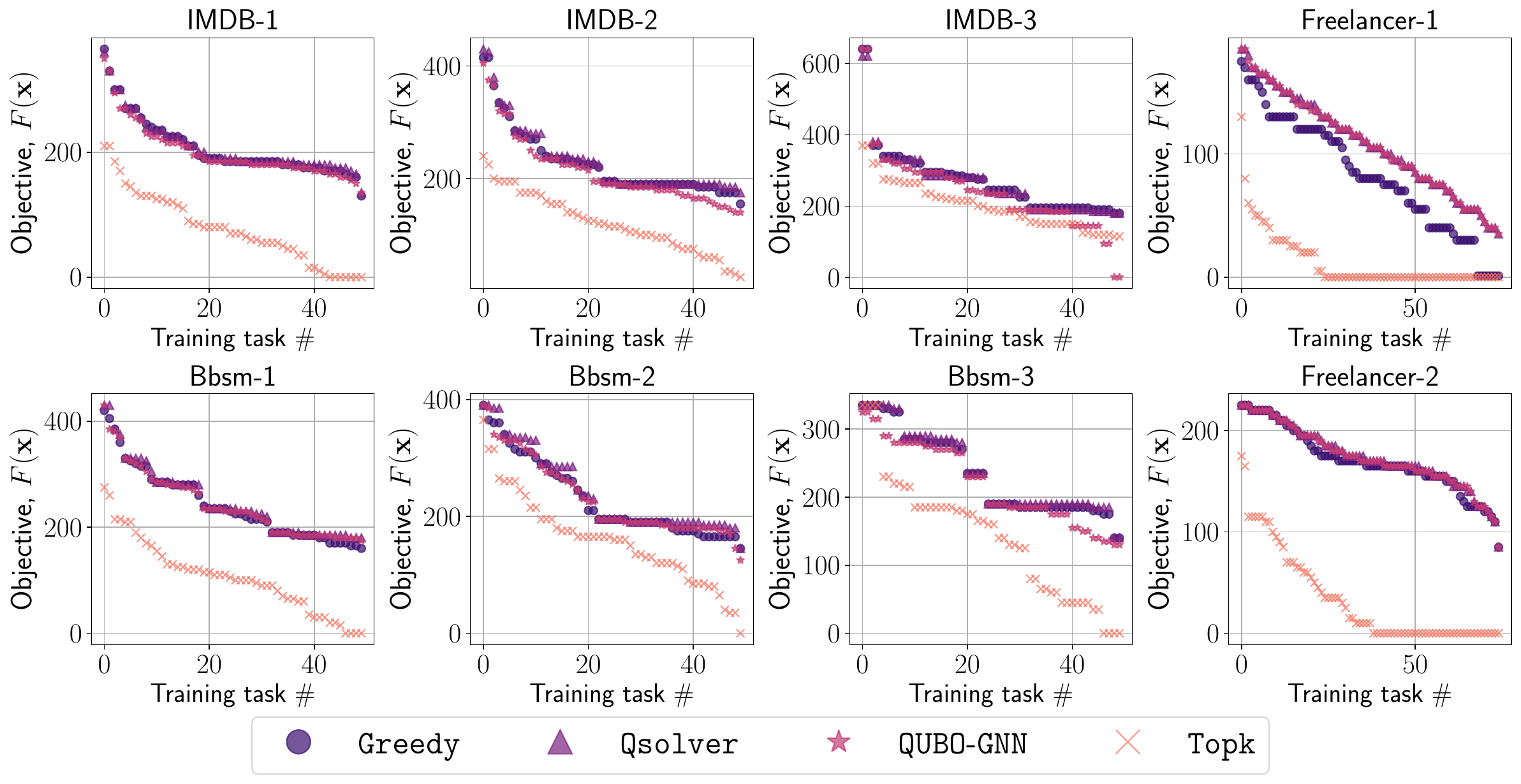}
    \caption{Comparative performance of {\qubognn}, {\qsolver}, {\greedy} and {\baselinetopk}, across individual training tasks, in terms of the sorted objective $\objective()$.}
    \label{fig:covCost-training}
\end{figure}
Figure~\ref{fig:covCost-training} presents a scatter plot of the objectives $\objective$ for each training task instance (for each dataset) for {\coveragelinearcost}; the tasks are sorted in decreasing order of $\objective$. We conclude that {\qubognn} is competitive with {\greedy} and even outperforms it in multiple cases, demonstrating that GNN-based approaches can achieve strong performance even without explicit heuristic tuning. 
Furthermore, for the {\freelancerone} dataset, both {\qubognn} and {\qsolver} outperform {\greedy} by over 30\%.
In our experiments, {\qubognn} consistently selects experts based on their skill relevance and almost never violates the constraints of the underlying LPs; thus {\qubognn} can identify well-balanced teams without the need for additional filtering mechanisms. 

\spara{Investigating Node Embeddings.}
\begin{figure}[ht]
    \centering
    \begin{subfigure}{0.46\textwidth}
    \includegraphics[width=\linewidth]{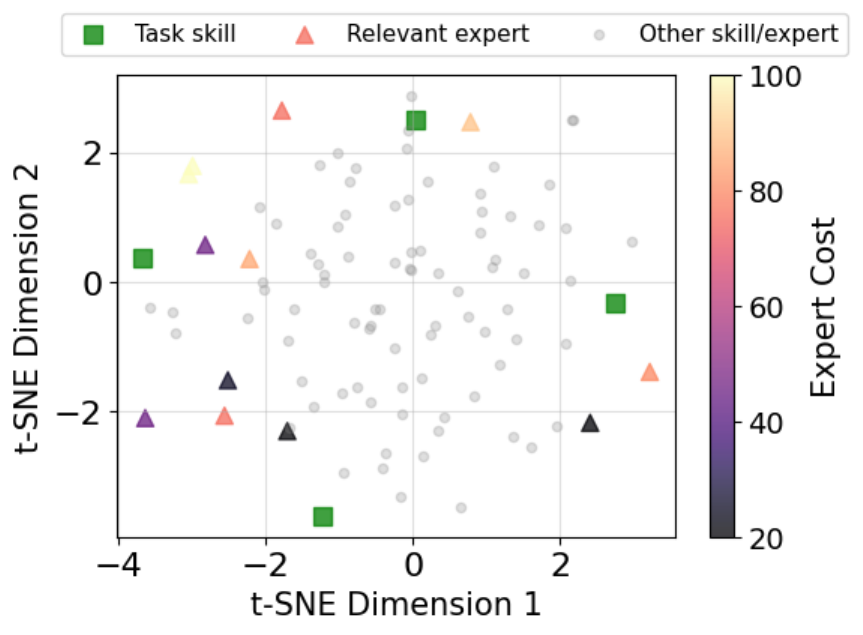}    
    \end{subfigure}
    \begin{subfigure}{0.46\textwidth}
    \includegraphics[width=\linewidth]{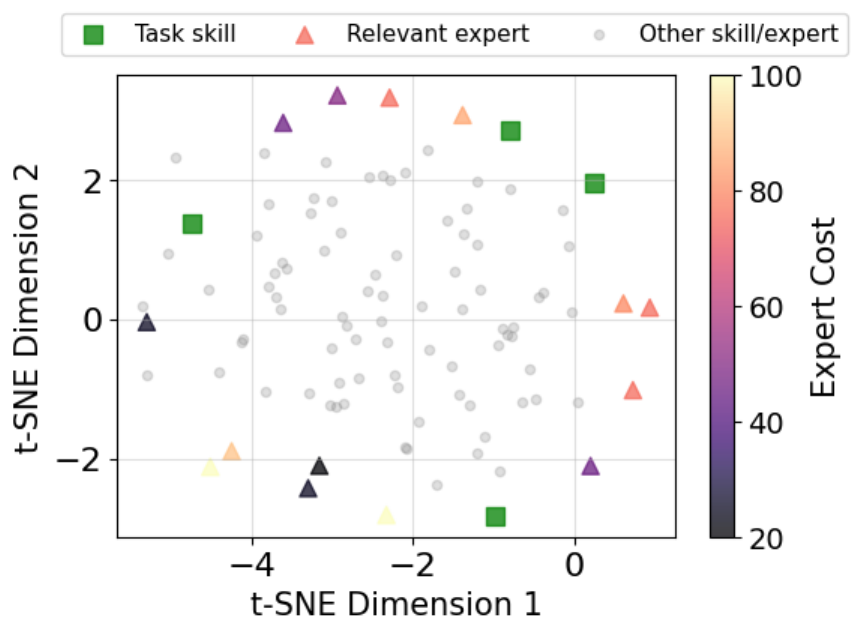}    
    \end{subfigure}
    \caption{Scatter plots of skill and expert node embeddings projected to 2D using t-SNE. Each set of embeddings was generated by a {\coveragelinearcost} {\qubognn} model after training on a task from {\freelancerone}.}
    \label{fig:node-embeddings}
\end{figure}
Figure~\ref{fig:node-embeddings} shows two scatter plots of skill and expert node embeddings projected to 2D using t-SNE~\cite{van2008visualizing}. Each set of embeddings was generated by a {\qubognn} model for {\coveragelinearcost} after training on a task from {\freelancerone}. This figure is representative of the patterns observed in node embeddings for all instances of {\teamformation}. We observe that the embeddings corresponding to task skills and relevant experts (i.e. experts who have at least one required skill) differentiate themselves from other skills/experts by forming an outer perimeter and occupying distinct regions of the plot. Experts with similar skills often cluster together, and their embeddings are often similar to those of their common skill(s). This indicates that {\qubognn} successfully learns representations between skills and experts and is able to correctly identify sets of experts that are important for covering a task.

\subsection{Transfer Learning}\label{sec:transfer-learning-exp}
Intuitively, we expect a {\qubognn} model $\mathcal{M}$ to learn node embeddings that result in good assignments for new tasks that are similar to the tasks $\mathcal{M}$ was trained on. 
Consider $\numtasks$ {\qubognn} models that have been trained on their corresponding tasks $\task_1, \ldots, \task_\numtasks$. Given an unseen task $\task '$, we first compute the Jaccard similarity of $\task'$ with each of $\task_1, \ldots, \task_\numtasks$, and select the {\qubognn} model $\mathcal{M}'$ corresponding to the task that is most similar to $\task'$. Next, we initialize the new {\teamformation} instance for $\task'$ with the pre-trained node embeddings corresponding to $\mathcal{M}'$, and use model $\mathcal{M}'$ to perform a single forward pass to obtain an assignment $\assignment$ for $\task'$. We refer to this method {\qubognnSimilar}. For each dataset, we evaluate it against the following two baselines on 100 new tasks.

\spara{{\qubognnRandom}:} We use a random sample of 3 pre-trained {\qubognn} models. We perform a single forward pass using each model and select the assignment $\assignment$ that yields the best objective.

\spara{{\qsolverSimilar}:} Given a new task $\task'$, we use the solution of {\qsolver} corresponding to the task (from $\task_1, \ldots, \task_\numtasks$) that has the highest Jaccard similarity to $\task'$ to compute the objective for $\task'$.

\begin{figure}[ht]
    \centering
    \includegraphics[width=0.95\textwidth]{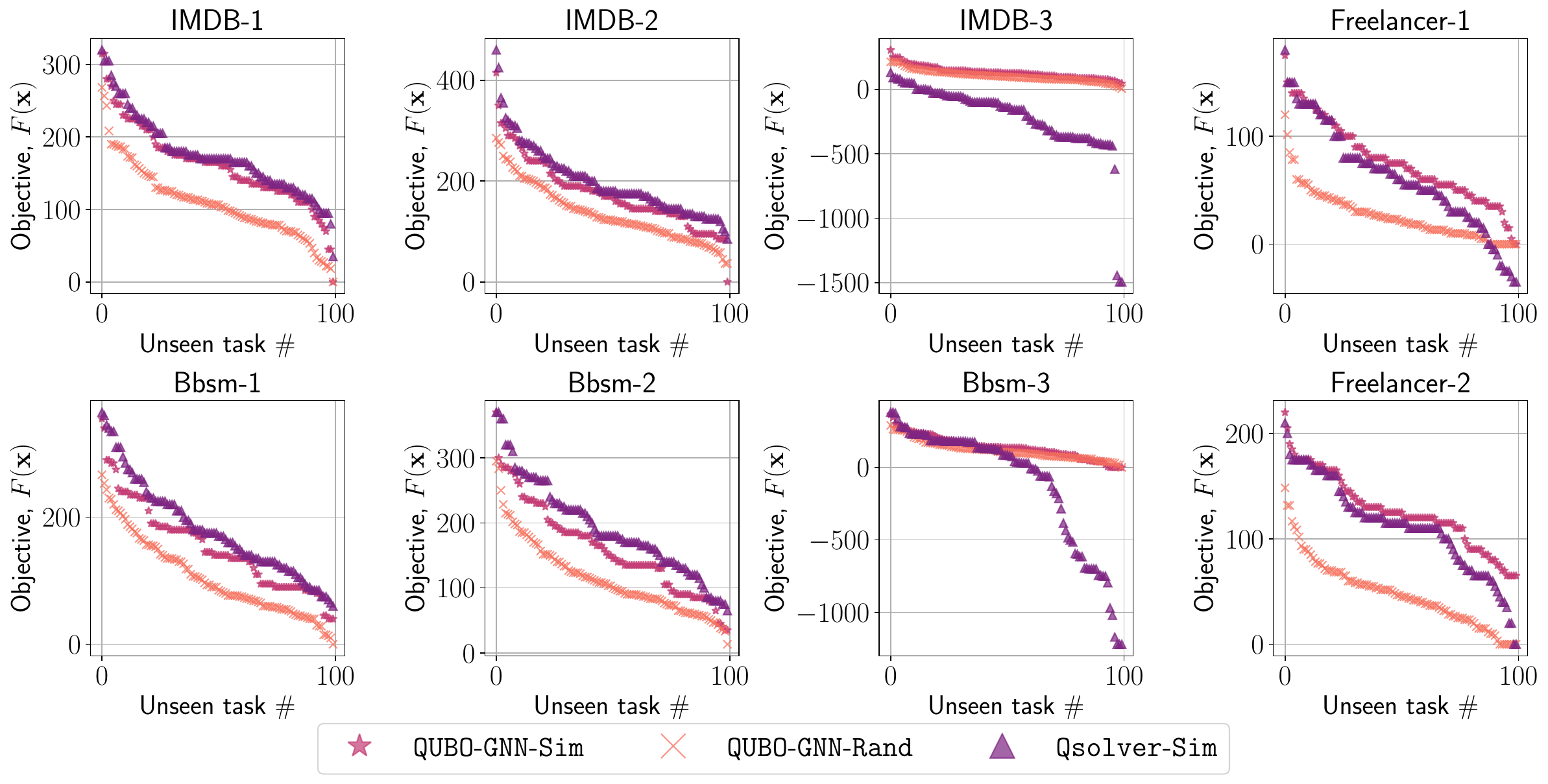}
    \caption{Evaluation of transfer learning on 100 new tasks across each dataset for {\coveragelinearcost} in terms of the sorted objective $\objective()$.}
    \label{fig:covCost-eval}
\end{figure}

Figure~\ref{fig:covCost-eval} shows a scatter plot of sorted objectives $\objective$ of {\qubognnSimilar} and the two baselines for 100 new tasks across each dataset for {\coveragelinearcost}. The results for the other two problems are shown in the supplement.

We note that {\qubognnSimilar} outperforms {\qsolverSimilar} for {\freelancer} and {\imdbthree} and {\bibsonomythree}, while the two methods have comparable performance for {\imdbone}, {\imdbtwo}, {\bibsonomyone} and {\bibsonomytwo}. {\qubognnRandom} has poor overall performance. This was expected, since using node embeddings of a random task would not necessarily aid solving of a new problem. {\qubognnSimilar} often finds solutions with high coverages, indicating that the learned node embeddings from the original model can capture useful relationships between skills and experts that can then be leveraged for other tasks. We also find, intuitively, that the efficacy of using node embeddings from a {\qubognn} model (trained on task $\task_i$) for a new task $\task_j$, correlates strongly with the Jaccard similarity of $\task_i$ and $\task_j$.

\section{Conclusions and Future Work}\label{sec:conclusions}
In this paper, we introduced a unified QUBO-based framework for the general {\teamformation} problem, enabling a versatile algorithmic approach across problem variants that balance task coverage with expert costs. 
We then evaluated our framework using both a QUBO solver, and a GNN method that maximizes the {\teamformation} objective by optimizing a QUBO-derived loss function.
In our experimental evaluation on real-world datasets from diverse domains, we demonstrated that our methods consistently find expert assignments with high objectives, often outperforming combinatorial baselines designed specifically for certain problem variants. Finally, we highlighted the potential for transfer learning, where learned representations from one problem instance can be effectively used to solve other related instances.

\spara{Future Work.} Finding optimal penalty parameters for our QUBO formulations is challenging, consequently opening up an avenue for future work on efficient methods to tune these penalties. A natural extension of our work could
consider multiple input tasks and explore more (complex) expert cost functions based on workload, team diameter, etc. Finally, there is scope for fine-tuning the {\qubognn} model architecture to improve performance.

\mpara{Acknowledgment.}
This work was supported by gifts from Microsoft and Google. 


\appendix  
\section*{Supplementary Material}

\section{QUBO Formulations for {\teamformation}}\label{sec:appendix-qubo}
For the derivations below, we use the vector $\vecy = \vecs \; || \; \assignment$ which represents the concatenation of $\vecs$ and $\assignment$, and the $(\numexperts \times \numskills)$ skill-membership matrix $E$. Given a length $n$ vector $\vecc$, we denote the $n \times n$ diagonal matrix with $\vecc$ along the main diagonal by $\text{diag}(\vecc)$.
We also use ${\bf e^{\numskills}_j}$, the standard basis vector of length $\numskills$ with 1 in the $j^{th}$ position and 0 elsewhere.

For any inequality constraint of the form \( h(\assignment ) = b - \sum_i a_i x_i \geq 0 \), we use the expansion of the exponential decay curve to quadratic order: $e^{-h(\assignment)} \approx 1 - h(\assignment) + \frac{1}{2}h(\assignment)^2$. 
Then we add a penalization term $\big(-p_1 h(\assignment) + p_2 h(\assignment)^2\big)$ to the objective, where $p_1, p_2$ are tunable penalty multipliers. 
Given a LP \(\min {\vecc}^T \assignment \) subject to \( A \assignment = \vecb \), we create penalty matrices $\penaltymat_1$ and $\penaltymat_2$ to capture the linear $(h(\assignment))$ and quadratic $(h(\assignment)^2)$ components of the penalization term respectively.
Then the corresponding $Q$-matrix is given by 
$$Q = \text{diag}(\vecc) - \penaltymat_1 + \penaltymat_2$$
Below we provide all mathematical details to use unbalanced penalization to formulate LPs corresponding to our {\teamformation} variants into QUBO.

\subsection{{\maxcover}}\label{appendix:maxcover-qubo}
The objective is to maximize $\objective(\assignment) = \lambda \cov(\task \mid \assignment)$ under the constraint $\costfunctionk(\assignment) \leq k$. We use the vector $\vecy = \vecs \; || \; \assignment$, and introduce a length $(\numskills + \numexperts)$ vector \\
${\vecc} = (c_{1}, \ldots, c_{m+n})$ such that: 
$$c_i = \begin{cases} 
\lambda & \text{if } i \leq \numskills \text{ and skill }i \in \task\\ 
0 & \text{otherwise}.
\end{cases}$$ 
We observe that $\objective(\assignment) = {\vecc}^T {\vecy}$, and specify {\maxcover} as the following constrained linear program:
\begin{align*}
    \text{maximize}~~  & {\vecc}^T {\vecy},\\
    \text{such that}~~ & \sum_{i=1}^{\numexperts} x_{i} \leq k \\
    &s_j - \sum_{i=1}^{\numexperts} E(i, j) \cdot x_i \leq 0 \quad \text{ for all } 1 \leq j \leq \numskills , \text{ and}\\
    & s_i, x_{i}\in \{0, 1\}.
\end{align*}
To encode the unbalanced penalization with tunable penalties $p_1, p_2$, we create penalty matrices corresponding to the LP constraints.
\begin{enumerate}
    \item \textit{Cardinality Constraint.} We can write the size constraint as $h_k(\assignment) = k - \sum_{i=1}^{\numexperts} x_{i} \geq 0$, and expand $e^{-h_k(\assignment)} = -p_1 h_k(\assignment) + p_2 h_k(\assignment)^2$, which gives:
    \begin{align*}
    e^{-h_k(x)} &= - p_1 \big(k - \sum_{i=1}^{\numexperts} x_{i}\big) + p_2 \big(k - \sum_{i=1}^{\numexperts} x_{i} \big)^2\\
            &= (p_1 - 2k p_2)\sum_{i=1}^{\numexperts} x_{i} + p_2 \big(\sum_{i=1}^{\numexperts} x_{i}\big)^2 + p_2 k^2 - p_1 k
    \end{align*}
    We denote the length $\numexperts$ vector of ones $\mathbf{1}_{\numexperts}$, and define $\hat{\penaltymat_k} = p_2 \big(2 \cdot \mathbf{1}_{\numexperts} \mathbf{1}_{\numexperts}^T - \text{diag}(\mathbf{1}_{\numexperts} \mathbf{1}_{\numexperts}^T)\big)$. Then we create the $(\numskills + \numexperts) \times (\numskills + \numexperts)$ penalty matrix $\penaltymat_k$ to capture both the linear and quadratic terms in $e^{-h_k(\assignment)}$ as follows:
    \begin{equation*}
        \penaltymat_k = (p_1 - 2k p_2)\cdot \text{diag}(\mathbf{0}_{\numskills} \; || \; \mathbf{1}_{\numexperts}) + 
        \begin{bmatrix}
        \mathbf{0}_{m \times m} & \mathbf{0}_{m \times n} \\
        \mathbf{0}_{n \times m} & \; \hat{\penaltymat_k}
        \end{bmatrix}
    \end{equation*}
    
    \item \textit{Coverage Constraint.} We note that $s_j - \sum_{i=1}^{\numexperts} E(i, j) \cdot x_i = y_j - \sum_{i=\numskills + 1}^{\numskills + \numexperts} E(i - \numskills, j) \cdot y_i$. Then for each $1 \leq j \leq \numskills$ we have the constraint $h_j(\vecy) = \sum_{i=\numskills + 1}^{\numskills + \numexperts} E(i - \numskills, j) \cdot y_i - y_j \geq 0$.  i.e. $e^{-h_j(\vecy)} \approx -p_1 h_j(\vecy) + p_2 h_j(\vecy)^2$, we denote the length $(\numskills + \numexperts)$ vector ${\bf \hat{h}_j} = \big({- \bf e^{\numskills}_j} \; ||\; (E_{1,j}, E_{2,j}, \ldots, E_{\numexperts,j})\big)$, and then use $\bf \hat{h}_j$ to construct the following matrix $\penaltymat_{C}$ corresponding to the linear and quadratic penalties for all $m$ coverage constraints.

    \begin{equation*}
    \penaltymat_{C} = -\text{diag}\Big(\sum_{j=1}^{\numskills} p_1 {\bf \hat{h}_j}\Big) + \sum_{j=1}^{\numskills} p_2 \big(2 {\bf \hat{h}_j} {\bf \hat{h}_j}^T - \text{diag}( {\bf \hat{h}_j} {\bf \hat{h}_j}^T)\big)
    \end{equation*}
\end{enumerate}
Then the $(\numskills + \numexperts) \times (\numskills + \numexperts)$ square matrix $Q = -\text{diag}({\vecc}) - \penaltymat_k + \penaltymat_C$ is a complete QUBO formulation of {\maxcover}, where minimizing ${\vecy}^T Q {\vecy}$ corresponds to maximizing $\objective(\assignment) = \lambda \cov(\task \mid \assignment) - \costfunctionk(\assignment)$.

\subsection{{\coveragelinearcost}}\label{appendix:linearcost-qubo}
The objective is to maximize $\objective(\assignment) = \lambda \cov(\task \mid \assignment)  - \costfunctionlinear (\assignment)$. We use the vector $\vecy = \vecs \; || \; \assignment$, and define the length $(\numskills + \numexperts)$ vector ${\vecc} = (c_{1}, \ldots, c_{(\numskills + \numexperts)})$ such that
$$c_i = \begin{cases} 
\lambda & \text{if } i \leq \numskills \text{ and skill } i \in \task\\ 
-\cost_{i-\numskills} & \text{if } i > \numskills\\
0 & \text{otherwise}.
\end{cases}$$ 
where $\cost_{i}$ corresponds to the cost of hiring expert $i$. Using this transform we have $\objective(\assignment) = {\vecc}^T {\vecy}$, and the following linear program encodes {\coveragelinearcost}:
\begin{align*}
    \text{maximize}~~  & {\vecc}^T {\vecy},\\
    \text{such that}~~ & s_j - \sum_{i=1}^{\numexperts} E(i, j) \cdot x_i \leq 0 \quad \text{ for all } 1 \leq j \leq \numskills , \text{ and}\\
    & s_i, x_{i}\in \{0, 1\}.
\end{align*}
\textit{Coverage Constraint.} We note that $s_j - \sum_{i=1}^{\numexperts} E(i, j) \cdot x_i = y_j - \sum_{i=\numskills + 1}^{\numskills + \numexperts} E(i - \numskills, j) \cdot y_i$.
Then, for each $1 \leq j \leq \numskills$ we have the constraint $h_j(\vecy) = \sum_{i=\numskills + 1}^{\numskills + \numexperts} E(i - \numskills, j) \cdot y_i - y_j \geq 0$. To encode the unbalanced penalization with tunable penalties $p_1, p_2$ i.e. $e^{-h_j(\vecy)} \approx -p_1 h_j(\vecy) + p_2 h_j(\vecy)^2$, we denote the length $(\numskills + \numexperts)$ vector ${\bf \hat{h}_j} = \big({- \bf e^{\numskills}_j} \; ||\; (E_{1,j}, E_{2,j}, \ldots, E_{\numexperts,j})\big)$, and then use $\bf \hat{h}_j$ to construct the following two matrices corresponding to the linear and quadratic penalties associated with all $m$ constraints.
\begin{itemize}
    \item $\penaltymat_1 = \text{diag}\Big(\sum_{j=1}^{\numskills} p_1 {\bf \hat{h}_j}\Big)$
    \item $\penaltymat_2 = \sum_{j=1}^{\numskills} p_2 \big(2 {\bf \hat{h}_j} {\bf \hat{h}_j}^T - \text{diag}( {\bf \hat{h}_j} {\bf \hat{h}_j}^T)\big)$
\end{itemize}
Then the $(\numskills + \numexperts) \times (\numskills + \numexperts)$ square matrix $Q = -\text{diag}({\vecc}) - \penaltymat_1 + \penaltymat_2$ provides a complete QUBO formulation of {\coveragelinearcost}, where minimizing ${\vecy}^T Q {\vecy}$ corresponds to maximizing $\objective(\assignment) = \lambda \cov(\task \mid \assignment)  - \costfunctionlinear (\assignment)$.

\subsection{{\coveragegraphcost}}\label{appendix:graphcost-qubo}
The objective is to maximize $\objective(\assignment) = \lambda \cov(\task \mid \assignment)  - \costfunctiongraph(\assignment)$, given a distance function $d(\cdot,\cdot)$ between any pair of experts. The following constrained LP encodes the {\coveragegraphcost} problem:

\begin{align*}
    \text{maximize}~~  & \lambda \cdot \sum_{i=1}^{n} s_i - \sum_{(i,j)} d(i,j) \cdot (x_i x_j)\\
    \text{such that}~~ &s_j - \sum_{i=1}^{\numexperts} E(i, j) \cdot x_i \leq 0 \quad \text{ for all } 1 \leq j \leq \numskills , \text{ and}\\
    & s_i, x_{i}\in \{0, 1\}.
\end{align*}
\textit{Coverage Constraint.} We note that $s_j - \sum_{i=1}^{\numexperts} E(i, j) \cdot x_i = y_j - \sum_{i=\numskills + 1}^{\numskills + \numexperts} E(i - \numskills, j) \cdot y_i$. Then, similar to the {\coveragelinearcost} problem, we apply unbalanced penalization with tunable penalties $p_1, p_2$ i.e. $e^{-h_j(\vecy)} \approx -p_1 h_j(\vecy) + p_2 h_j(\vecy)^2$, using the vector ${\bf \hat{h}_j} = \big({- \bf e^{\numskills}_j} \; ||\; (E_{1,j}, E_{2,j}, \ldots, E_{\numexperts,j})\big)$ to construct the linear and quadratic penalty matrices associated with all $m$ constraints: 
\begin{itemize}
    \item $\penaltymat_1 = \text{diag}\Big(\sum_{j=1}^{\numskills} p_1 {\bf \hat{h}_j}\Big)$ 
    \item $\penaltymat_2 = \sum_{j=1}^{\numskills} p_2 \big(2 {\bf \hat{h}_j} {\bf \hat{h}_j}^T -       
           \text{diag}( {\bf \hat{h}_j} {\bf \hat{h}_j}^T)\big)$
\end{itemize}
We define a length $(\numskills + \numexperts)$ vector ${\vecc} = (c_{1}, \ldots, c_{(\numskills + \numexperts)})$, and set 
$$c_i = \begin{cases} 
\lambda & \text{if } i \leq \numskills \text{ and skill } i \in \task\\ 
0 & \text{otherwise}.
\end{cases}$$ 
Then we compute the $(\numexperts \times \numexperts)$ matrix $D$ of pairwise distances such that $D(i,j) = d(\expert_i, \expert_j)$ and add it to the lower-right submatrix of $\text{diag}({\vecc})$ to obtain $\hat{D}$.
\begin{equation*}
 \hat{D} = \text{diag}({\vecc}) + 
        \begin{bmatrix}
        \mathbf{0}_{m \times m} & \mathbf{0}_{m \times n} \\
        \mathbf{0}_{n \times m} & \;D_{n \times n}
        \end{bmatrix}
\end{equation*}
Observe that $\objective(\assignment) = \vecy^T \hat{D} \vecy$ encodes the {\coveragegraphcost} objective. The $(\numskills + \numexperts) \times (\numskills + \numexperts)$ square matrix $Q = -\hat{D} - \penaltymat_1 + \penaltymat_2$ provides a complete QUBO formulation of {\coveragegraphcost}, where minimizing ${\vecy}^T Q {\vecy}$ corresponds to maximizing $\objective(\assignment) = \lambda \cov(\task \mid \assignment)  - \costfunctiongraph(\assignment)$.

\section{Additional Experimental Details}\label{sec:appendix-results}
\subsection{Descriptions of Experimental Datasets}\label{sec:appendix-datasets}
\mpara{{\freelancer}:} This dataset consists of random samples of real jobs that are posted by users online, and a random sample of real freelancers~\footnote{https://www.freelancer.com}.
The data consists of 993 jobs ({\ie}, tasks) that require certain discrete skills, and 1212 freelancers ({\ie}, experts) who possess skills. 

From the subset of experts that have at least 2 skills, we randomly sample 50 and 150 experts and 250 skills in order to form the set of experts in the 2 datasets, {\freelancerone} and {\freelancertwo}. We randomly sample 250 tasks to form the set of tasks for each dataset.

We form the coordination cost graph among the experts as follows:
there is an edge between any two experts and the cost of this edge is the Jaccard distance (one minus the Jaccard similarity) between the experts' skill sets. 
We note that the Jaccard distance takes values between 0 and 1, and is 0 if two experts have identical skill sets, and 1 if their skills are mutually exclusive.

\mpara{IMDB:} The data is obtained from the International Movie Database~\footnote{https://www.imdb.com/interfaces/}. The original dataset contains information about movies, TV shows and documentaries along with the associated actors, directors, movie year and movie genre(s). 
We simulate a team-formation setting where 
the movie genres correspond to skills, movie directors to experts, and actors to tasks. 
The set of skills possessed by a director or an actor is the union of genres of the movies they have participated in.
As an illustrative example, the director \textit{Christopher Nolan} has the skills \textit{\{drama, action, history, biography, sci-fi, thriller\}} and the actor \textit{Emma Stone} has the skills \textit{\{crime, comedy, sci-fi, animation, romance, adventure\}}. 
We create three data instances of different sizes by selecting all movies created since 2020, 2018 and 2015 and select actors and directors associated with at least two genres. 
For the directors, we make sure that each director has at least one actor in common with at least one other director.
We randomly sample 200, 400 and 1000 directors and 300 actors, to form 
the three datasets: {\imdbone}, {\imdbtwo} and {\imdbthree} respectively. 
We also sample 300 actors in each dataset, to form the set of tasks.

We create a social graph among the directors (experts). We form an edge between two directors if they have directed at least two common actors. The cost of the edge is set to $e^{-fD}\in (0,1)$, where $D$ is the number of common actors among the two directors. The weight decreases as the number of common actors $D$ between two directors increases. We set $f = \frac{1}{10}$ [2], getting a reasonable edge-weight distribution of coordination costs in the social graph.

\spara{Bibsonomy:} 
This dataset comes from a social bookmark and publication sharing system [5].  
Each publication is associated with a set of \emph{tags}; we filter tags for stopwords and use the 75 most common tags as skills.
We simulate a setting where certain prolific authors (experts) conduct interviews for other less prolific authors (tasks). 
An author's skills are the union of the tags associated with their publications and focus on authors with at least 12 papers.
We create three datasets by selecting all publications since 2020, 2015, and 2010, and select all authors who have at least two tags and have at least one paper in common with at least one other author.  
We randomly sample 250, 500, and 1000 prolific authors to form the set of experts in datasets {\bibsonomyone}, {\bibsonomytwo}, and {\bibsonomythree}. We randomly sample authors with less than 12 papers, to form the set of tasks.

We create a social graph among authors using co-authorship to define the strength of social connection; 
two authors are connected with an edge if they have written at least one paper together. Again the cost of the edge is set to $e^{-fD}$, where $D$ is the number of distinct co-authored papers.
Again we set $f=\frac{1}{10}$
and obtain a reasonable distribution of edge-weights in the graph.

\subsection{{\qubognn} Implementation Parameters}\label{sec:exp-hyperparams}
Table~\ref{tab:model-params} specifies the model parameters used to formulate the three {\teamformation} variants into QUBO and instantiate {\qubognn} to find solutions. Note that for all experiments we used $\lambda = 50$. For the QUBO penalty parameters $p_1, p_2$, we grid-searched over the heuristic range $[10^{-1}, 10^2]$ for each training task, and selected the values $p_1, p_2$ that yielded the best task-specific objectives.

For embedding dimensions ($d_0, d_h$) we report values in terms of the number of skills $\numskills$ and experts $\numexperts$, which are dataset-dependent. We allow the {\qubognn} model to train for up to $\sim 10^5$ epochs, with a simple early stopping rule set to an absolute tolerance of $10^{-3}$ and a patience of $3 \times 10^3$.

\begin{table}[ht]
    \centering
    \caption{{\qubognn} model parameters for each variant for {\teamformation}.}
    \label{tab:model-params}
    \begin{tabular}{p{4cm} p{2cm} p{2cm} p{1.2cm} p{1.2cm} p{1.2cm} }
    \toprule
    {\teamformation} Variant & $d_0$ & $d_h$ & $p_d$ & $\alpha$ & $\beta$\\
    \midrule
        {\maxcover} & $(\numskills + \numexperts)/2$ & $(\numskills + \numexperts)/4$& 0.25 &2 & $10^{-3}$\\
        {\coveragelinearcost} & $(\numskills + \numexperts)/2$ & $(\numskills + \numexperts)/4$& 0.2 &1.5 & $5\times10^{-3}$\\
        {\coveragegraphcost} & $(\numskills + \numexperts)/2$ & $(\numskills + \numexperts)/4$& 0.25 &2.5 & $10^{-2}$ \\
    \bottomrule
    \end{tabular}
\end{table}

\subsection{Experimental Results for {\maxcover}}
We evaluate our algorithms against the baselines with respect to our overall objective.
We observe that {\qsolver} has the best performance for all datasets for {\maxcover}. We use $\meancov = \frac{1}{\numtasks}\sum_{i=1}^t \cov(\task_i | \assignment)$ to denote the mean coverage, and $\meansize = \frac{1}{\numtasks}\sum_{i=1}^t z_i$ to denote the mean solution size, across training tasks $\task_1, \ldots, \task_\numtasks$.
We observe from Table~\ref{tab:maxCov-Train} that all three methods find solutions yielding high coverages but {\qsolver} and {\greedy} both find the highest values of $\meancov$. 

\begin{table}[H]
\caption{Mean task coverage, $\meancov$ and solution size, $\meansize$ of {\qubognn}, {\qsolver} and {\greedy} across all training task instances for {\maxcover} with $k=3$.}
\centering
\label{tab:maxCov-Train}
\begin{tabular}{l *{3}{>{\centering\arraybackslash}m{1.5cm}} *{3}{>{\centering\arraybackslash}m{1.5cm}}}
\toprule
Dataset & \multicolumn{3}{c}{Mean Task Coverage} & \multicolumn{3}{c}{Mean Solution Size} \\
\cmidrule(r){2-4} \cmidrule(l){5-7}
& {\qubognn} & {\qsolver} & {\greedy} & {\qubognn} & {\qsolver} & {\greedy} \\
\midrule
{\freelancerone} & 0.75 & 0.89 & 0.88 & 2.0 & 2.7 & 2.6 \\
{\freelancertwo} & 0.80 & 0.99 & 0.97 & 2.3 & 2.8 & 2.7 \\
{\imdbone}     & 0.94 & 0.99 & 1.00 & 1.7 & 2.6 & 1.8 \\
{\imdbtwo}    & 0.87 & 1.00 & 1.00 & 1.5 & 2.5 & 1.4 \\
{\imdbthree}     & 0.77 & 1.00 & 1.00 & 1.2 & 2.2 & 1.1 \\
{\bibsonomyone}    & 0.95 & 1.00 & 1.00 & 1.6 & 2.6 & 1.6 \\
{\bibsonomytwo}    & 0.86 & 1.00 & 1.00 & 1.2 & 2.4 & 1.4 \\
{\bibsonomythree}   & 0.82 & 1.00 & 1.00 & 1.3 & 3.1 & 1.3 \\
\bottomrule
\end{tabular}
\end{table}

We note that {\qsolver} often find assignments with a larger solution size than {\greedy}, but the size of these assignments almost always satisfy the $k$-cardinality constraint; we observed a few violations for {\bibsonomythree}. {\qubognn} also satisfies the $k$-cardinality constraint but performs between 5-20\% worse than {\qsolver}. This gap in performance could be attributed to insufficient hyperparameter tuning, or suboptimal constraint penalization, since the {\qsolver} solution indicates that the QUBO formulation is able to find an assignment of experts that balances coverage with the constraints for {\maxcover}.

\begin{figure}[H]
    \includegraphics[width=\textwidth]{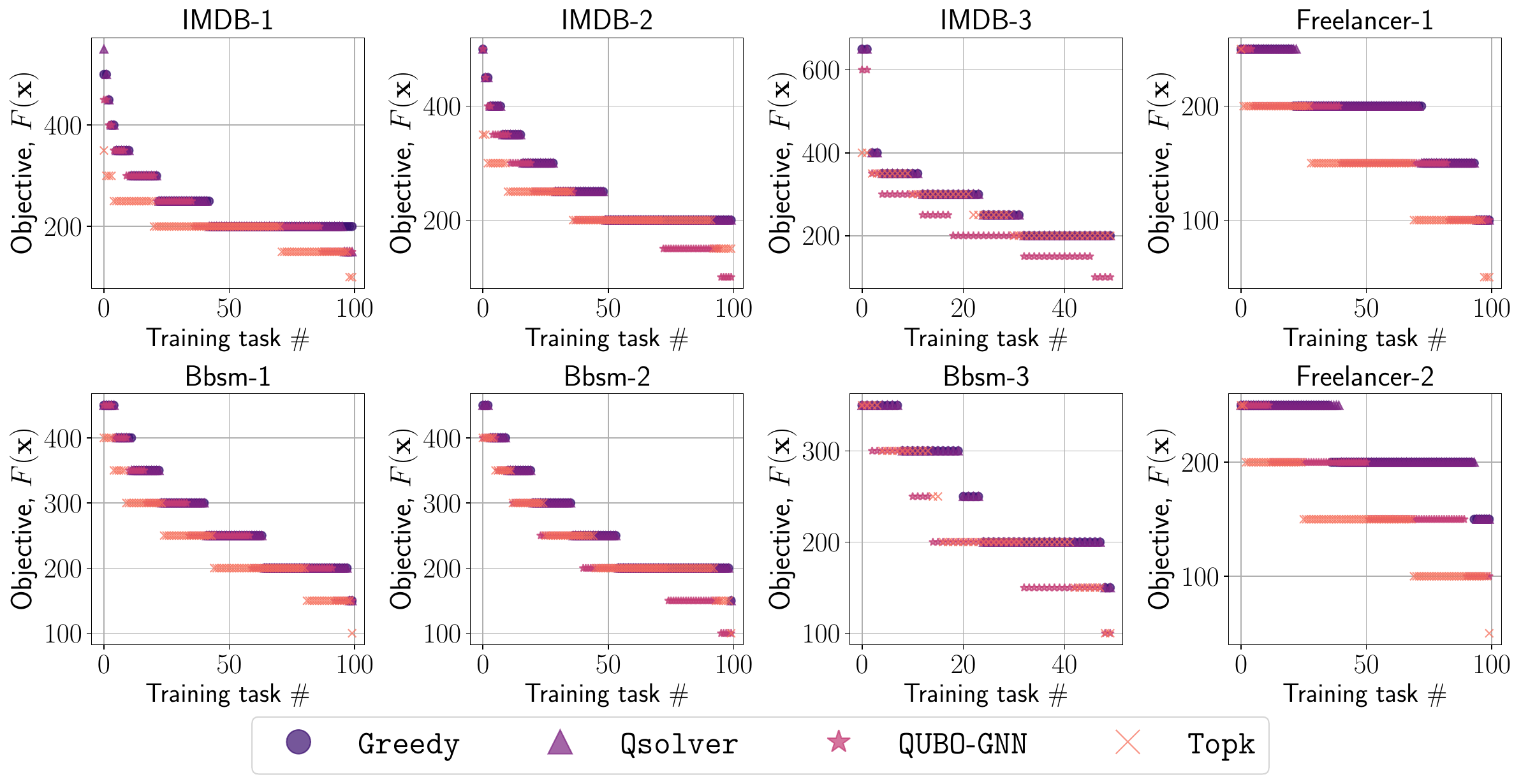}
    \caption{Comparative performance of {\qubognn}, {\qsolver}, {\greedy} and {\baselinetopk}, across individual training tasks, in terms of the sorted objective $\objective()$ for {\maxcover} with $k=3$.}
    \label{fig:maxCover-training}
\end{figure}

Figure~\ref{fig:maxCover-training} presents a scatter plot of the objectives $\objective$ for each training task instance (for each dataset) for {\maxcover}; the tasks are sorted in decreasing order of $\objective$. We observe that while {\qsolver} has the best performance, {\qubognn} is also competitive with {\greedy} for several training task instances. We observe that while {\baselinetopk} finds solutions with the lowest objective values, it doesn't perform very poorly. Since we set $\lambda = 50$, the objective values for {\maxcover} are all multiples of 50, and Fig.~\ref{fig:maxCover-training} illustrates the competitive (overlapping) objectives found by {\qubognn}, {\qsolver} and {\greedy} across the datasets.

\subsection{Experimental Results for {\coveragegraphcost}}
{\qsolver} has the best performance for all datasets for {\coveragegraphcost}.
We observe from Table~\ref{tab:CovEdges-Train} that all three methods find solutions yielding high coverages but {\qsolver} and {\greedy} both find the highest values of $\meancov$, while {\qubognn} finds solutions with slightly lower $\meancov$. 

\begin{table}[H]
\caption{Mean task coverage, $\meancov$ and solution size, $\meansize$ of {\qubognn}, {\qsolver} and {\greedy} across all training task instances for {\coveragegraphcost}.}
\centering
\label{tab:CovEdges-Train}
\begin{tabular}{l *{3}{>{\centering\arraybackslash}m{1.5cm}} *{3}{>{\centering\arraybackslash}m{1.5cm}}}
\toprule
Dataset & \multicolumn{3}{c}{Mean Task Coverage, $\meancov$} & \multicolumn{3}{c}{Mean Solution Size, $\meancov$} \\
\cmidrule(r){2-4} \cmidrule(l){5-7}
& {\qubognn} & {\qsolver} & {\greedy} & {\qubognn} & {\qsolver} & {\greedy} \\
\midrule
{\freelancerone} & 0.54 & 0.66 & 0.62 & 1.4 & 1.8 & 1.6 \\
{\freelancertwo} & 0.59 & 0.78 & 0.73 & 1.3 & 2.1 & 1.8 \\
{\imdbone}     & 0.75 & 0.89 & 0.85 & 1.1 & 1.3 & 1.2 \\
{\imdbtwo}    & 0.86 & 0.97 & 0.95 & 1.0 & 1.2 & 1.1 \\
{\imdbthree}     &  0.71 & 0.97 & 0.98 & 1.1 & 1.8 & 1.0 \\
{\bibsonomyone}    & 0.90 & 0.95 & 0.95 & 1.2 & 1.4 & 1.4 \\
{\bibsonomytwo}    & 0.90 & 0.99 & 0.97 & 1.0 & 1.2 & 1.1 \\
{\bibsonomythree}   & 0.86 & 1.00 & 0.99 & 1.1 & 1.3 & 1.2 \\
\bottomrule
\end{tabular}
\end{table}

\begin{figure}[H]
    \centering
    \includegraphics[width=0.9\textwidth]{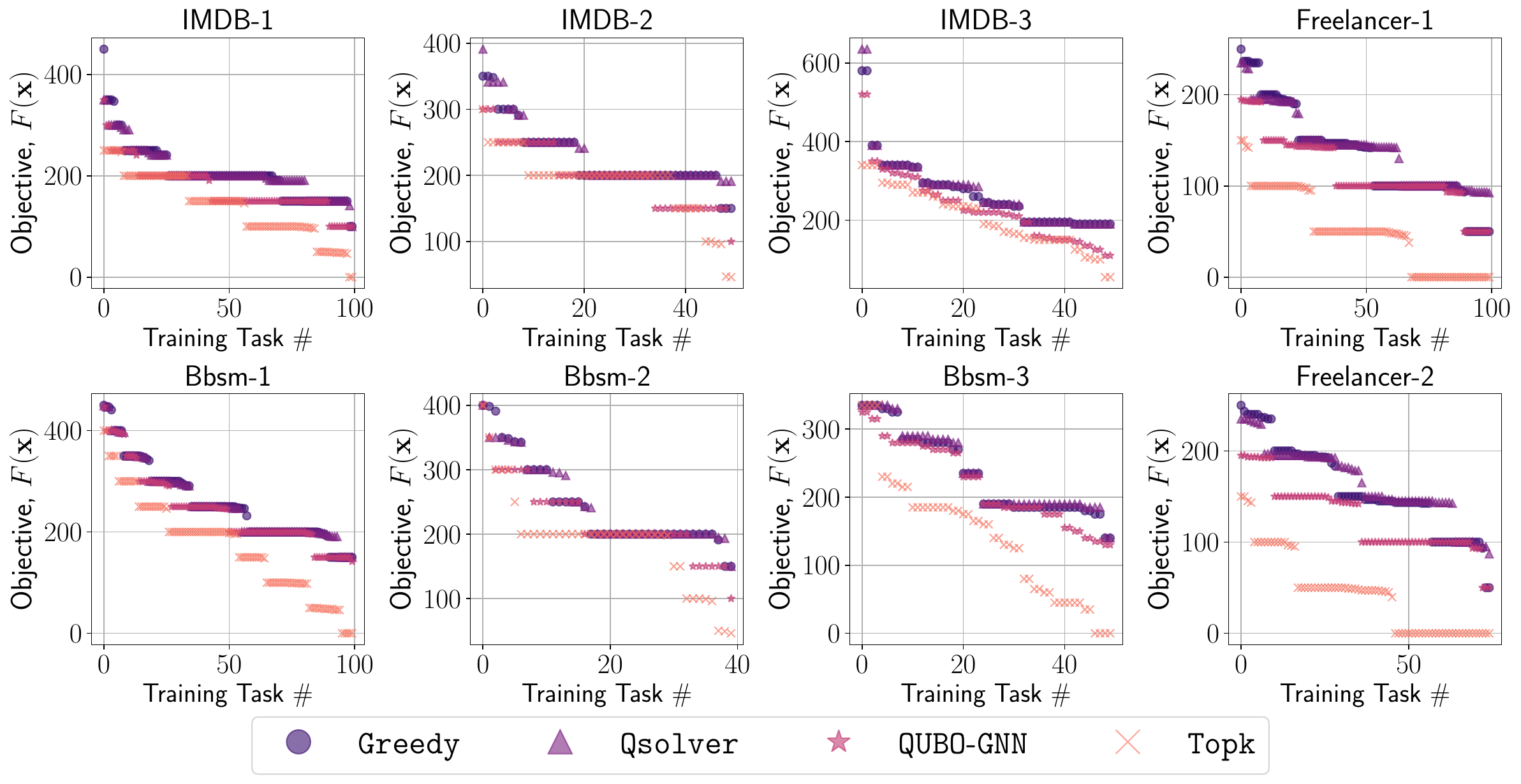}
    \caption{Comparative performance of {\qubognn}, {\qsolver}, {\greedy} and {\baselinetopk}, across individual training tasks, in terms of the sorted objective $\objective()$ for {\coveragegraphcost}.}
    \label{fig:covEdges-training}
\end{figure}

We note again that {\qsolver} often find assignments with a larger solution size than {\greedy}, but {\qubognn} does not.
Figure~\ref{fig:covEdges-training} presents a scatter plot of the objectives $\objective$ for each training task instance for {\coveragegraphcost}; the tasks are sorted in decreasing order of $\objective$. 
We observe that {\qubognn} is competitive within 10-20\% of {\qsolver} and {\greedy} for all datasets.

Both {\qsolver} and {\qubognn} are successfully able to find assignments that select a small number of -- typically less than three -- relevant experts from a much larger set of experts. Investigating individual assignments, we also observe that {\qsolver} and {\qubognn} often picked experts with complimentary sets of skills that had a large intersection of skills with the task.

Since the cost function $\costfunctiongraph$ depends on the pairwise structure of the graph, we observe that our {\qubognn} model is able to learn representations of skills and experts that capture this structure. While there is scope for improvement via further penalty parameter and hyperparameter tuning, we observed that for certain task instances (in different datasets) {\qubognn} performed on par with {\qsolver}.

\subsection{Transfer Learning: {\maxcover} \& {\coveragegraphcost}}
\label{sec:appendix-results-transfer}

\begin{figure}[ht]
    \centering
    \includegraphics[width=0.84\textwidth]{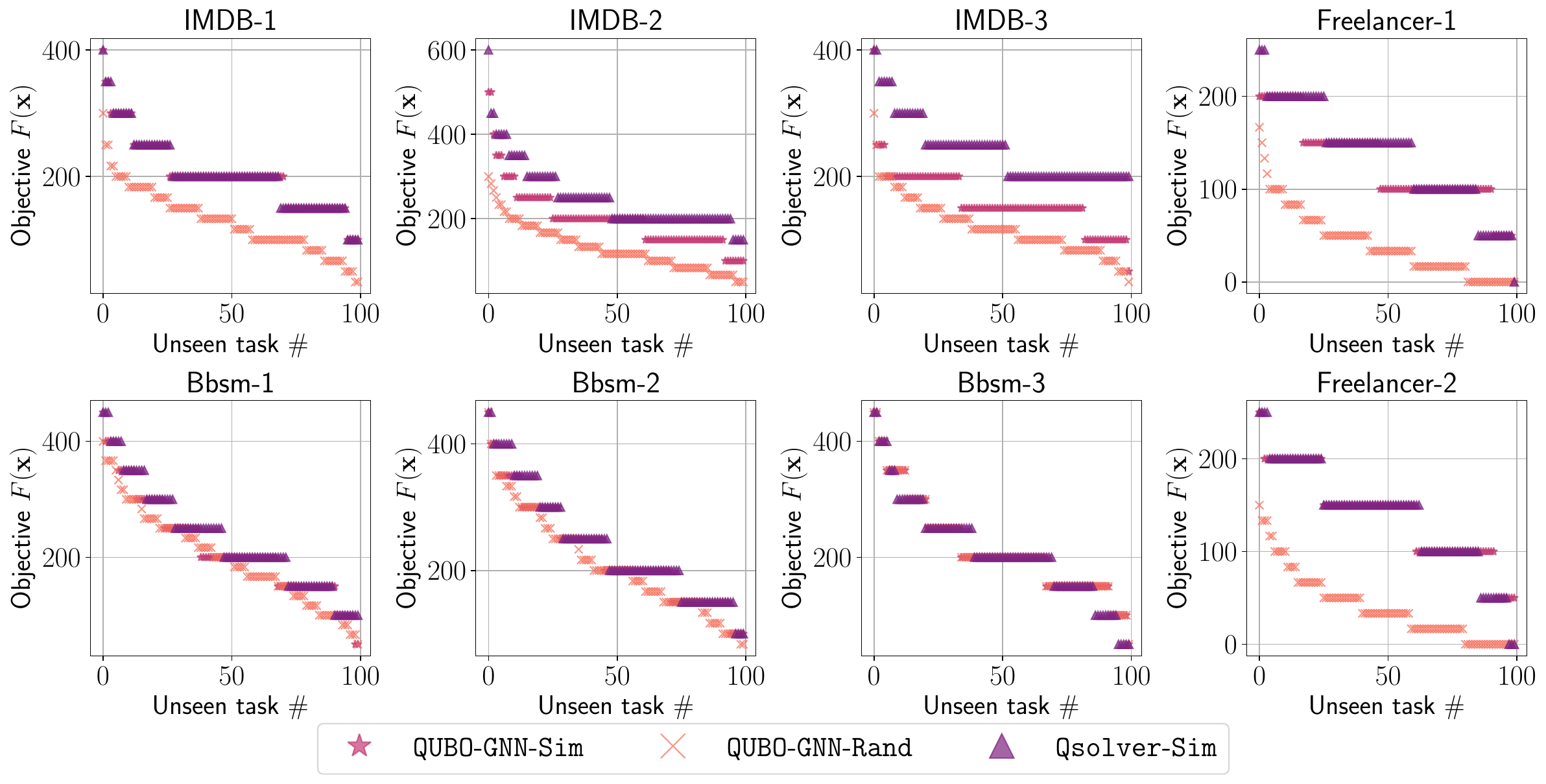}    \caption{Evaluation of transfer learning on 100 new tasks across each dataset for {\maxcover} in terms of the sorted objective $\objective()$.}
    \label{fig:maxCover-eval}
\end{figure}

\begin{figure}[ht]
    \centering
    \includegraphics[width=0.83\textwidth]{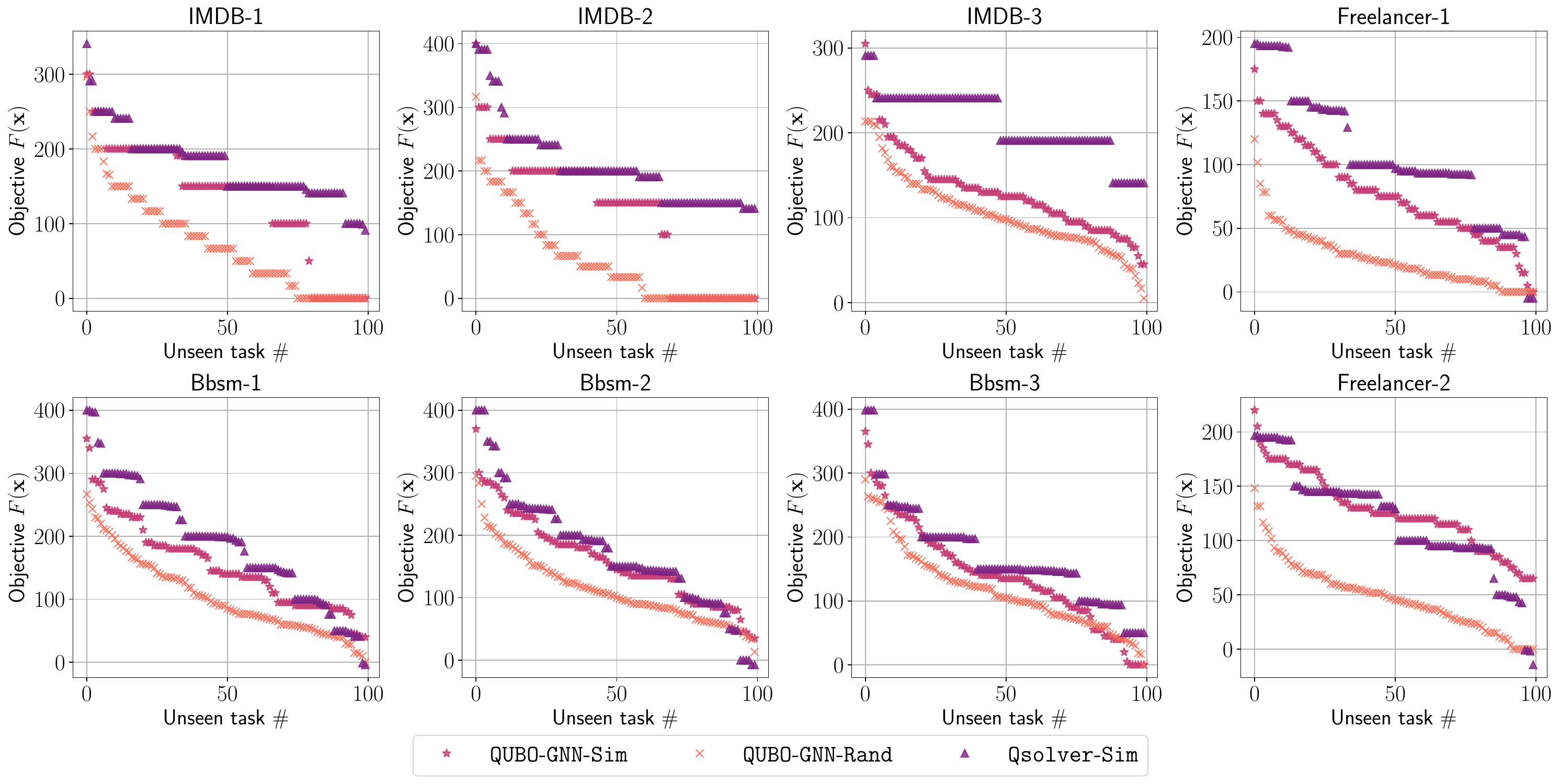}
    \caption{Evaluation of transfer learning on 100 new tasks across each dataset for {\coveragegraphcost} in terms of the sorted objective $\objective()$.}
    \label{fig:covEdges-eval}
\end{figure}

From Figures~\ref{fig:maxCover-eval} and~\ref{fig:covEdges-eval} we observe that {\qubognnSimilar} and {\qsolverSimilar} have comparable performance on most datasets except {\imdbthree}. {\qubognnSimilar} often finds solutions with high objectives: thus learned node embeddings from the original {\qubognn} model can capture useful relationships between skills and experts. We observed that in general when the Jaccard similarity between two tasks was high, so was the efficacy of using the node embeddings from the pre-trained model for transfer learning.

We observe that {\qubognnRandom} performs poorly on most datasets, again confirming our hypothesis that a model trained on a specific task would be unlikely to be helpful to solve a team formation problem for a random (unrelated) task. For {\coveragegraphcost} we see observe that for certain datasets {\qsolverSimilar} has superior performance, indicating that transfer learning requires problems that yield pairwise task similarities that are conducive to re-using pre-trained node embeddings.

\end{document}